\documentclass[]{mosi}

\usepackage{array}
\usepackage{makecell}
\usepackage{threeparttable}
\usepackage{times}
\usepackage{microtype}
\usepackage{amsmath,amssymb,mathtools}
\usepackage{booktabs,tabularx,multirow}
\usepackage{enumitem}
\usepackage{needspace}
\usepackage{fvextra}
\usepackage{xcolor}
\usepackage{tikz}
\usepackage{xspace}
\usepackage{url}

\newcommand{\lgw}{\texorpdfstring{\textup{LatentGUIWorld}}{LatentGUIWorld}\xspace}
\newcommand{\move}{\texorpdfstring{\textsc{Move}}{MOVE}\xspace}

\title{Learn from Move: the Next Step for GUI Agents}

\renewcommand{\authorformat}[2][]{%
  \mbox{{\authorfont\bfseries #2\ifstrempty{#1}{}{\textsuperscript{#1}}}}%
}
\renewcommand{\authorlist}{%
  \authorformat[1,2,*]{Jiahao Lu}\quad
  \authorformat[3,2,*]{Yize Shen}\quad
  \authorformat[3]{Chuanyuan Tan}\quad
  \authorformat[1]{Mianqiu Huang}\par
  \vspace{1pt}%
  \authorformat[1,2]{Shicheng Fang}\quad
  \authorformat[1]{Xinchi Chen}\quad
  \authorformat[1,2]{Yining Zheng}\quad
  \authorformat[2,$\dagger$]{Yuxin Wang}\quad
  \authorformat[1,2,$\dagger$]{Xipeng Qiu}\par
  \vspace{2.5mm}%
  {\normalfont\fontsize{11}{14}\selectfont
    \textsuperscript{1}Fudan University\par
    \textsuperscript{2}Shanghai Innovation Institute\par
    \textsuperscript{3}Soochow University\par
  }%
  \vspace{1.5mm}%
  {\normalfont\small \textsuperscript{*}Equal contribution.\quad
    \textsuperscript{$\dagger$}Corresponding authors.\par}%
}

\abstract{
Existing benchmarks for GUI agents primarily focus on increasing task difficulty by extending task horizons. Besides, enabling agents to explore unfamiliar environments and successfully complete tasks within them is also a critical yet challenging problem. We introduce \lgw, a benchmark of six tasks in three matched pairs organized by the evidence needed for control: $I_0$ uses prior knowledge and available observations, $I_1$ requires interaction that directly reveals hidden semantics, and $I_2$ identifies dynamics from action--response relationships. Frontier models struggle especially on $I_2$ tasks. We introduce \move, which separates movement from button events and exposes intermediate visual feedback, together with demonstrations of iterative exploration and control. Ablations show that both action granularity and multi-step demonstrations improve performance. After SFT, LatentLearner-9B achieves 50.0\% overall success on \lgw, comparable to several frontier models. Joint RL further raises egocentric Drag and Ten-Choice success to 92.0\% and 75.3\%. Behavioral analysis reveals multi-step exploration. In egocentric Drag, RL increases Pearson correlation between error-normalized commands and ideal gain compensation from 0.610 to 0.803. Together, these behaviors show that models learn to explore through interaction and adapt their control to environment dynamics. The same format improves GUI grounding after RL, reaching 67.37\% mean accuracy across five benchmarks versus 66.72\% for conventional RL.
}

\checkdata[Correspondence]{\href{mailto:lujiahao@sii.edu.cn}{lujiahao@sii.edu.cn}, \href{mailto:wangyuxin@sii.edu.cn}{wangyuxin@sii.edu.cn}, \href{mailto:xpqiu@fudan.edu.cn}{xpqiu@fudan.edu.cn}}
\checkdata[Repository]{\url{https://github.com/ti-mm/LearnFromMove}}
\checkdata[Benchmark]{\url{https://huggingface.co/datasets/OpenMOSS-Team/LatentGUIWorld}}

\begin{document}
\maketitle

\section{Introduction}

Multimodal agents combine visual perception, language understanding, and action to assist with a broad range of tasks. Computer use is a practical setting for these capabilities, spanning web browsing, document editing, and workflows across applications~\cite{xie2024osworld}. GUI agents carry out such tasks by interpreting screenshots and operating graphical user interfaces through mouse and keyboard actions, using the same interfaces available to human users.

As GUI agents tackle more complex tasks, benchmarks increasingly emphasize longer workflows and sustained execution across applications~\cite{xie2024osworld,yuan2026osworld2}. Yet completing long sequences of screen-based actions remains costly and unreliable~\cite{song2025coact1,zhou2026guivscli}. These difficulties have motivated hybrid agents that combine GUI interaction with command-line execution, code, or application tools. CoAct-1 and OSWorld-MCP show how programmatic actions can reduce reliance on lengthy GUI sequences and improve task completion~\cite{song2025coact1,jia2025osworldmcp}. A matched GUI--CLI study likewise finds that CLI agents with verifier-guided skill augmentation outperform screen-only GUI agents~\cite{zhou2026guivscli}. This shift motivates evaluating visual interaction beyond workflow length: what can an agent learn by acting on an unfamiliar interface and observing its response?

Coding agents investigate unfamiliar repositories by reading code, running programs, and using test feedback to guide subsequent changes~\cite{yang2024sweagent}. GUI agents face an analogous need when the correct action depends on properties that are absent from the current screenshot. Consider a rotation task whose response direction and sensitivity vary across episodes. An agent may know the goal and possible control rules, yet still need to move, observe the resulting rotation, and infer how to complete the task. Existing benchmarks include information acquisition and adaptation, but seldom isolate the evidence required to identify such episode-specific properties. We therefore study whether GUI agents can identify these hidden properties from action-induced visual feedback and use what they learn to guide subsequent control.

We introduce \lgw, a multimodal benchmark of six short GUI tasks organized into three matched pairs. Its taxonomy classifies tasks by whether sufficient knowledge is available before a task-directed action, and by how missing knowledge must be acquired: $I_0$ uses prior knowledge and observed context; $I_1$ requires a probe outcome that directly reveals hidden semantics; and $I_2$ requires interpreting a visual transition together with the action that caused it to identify hidden dynamics. The benchmark pairs exocentric Drag ($I_0$) with egocentric Drag ($I_2$), exocentric Ten-Choice ($I_1$) with its egocentric variant ($I_2$), and two Rotation geometries ($I_2$--$I_2$). These pairs preserve task objectives while varying the interaction mechanism, making environment identification an explicit difficulty dimension alongside task horizon.

\begin{figure*}[t]
\centering
\input{figures/overview-artwork.tex}
\caption{\textbf{Measure identification; expose evidence before commitment.}
\lgw spans three task pairs and evidence requirements $I_0$--$I_2$.
The rotation example relates an executed move to its visual consequence to identify
an unknown response gain and adapt control. The \move interface returns an
observation for a new decision after each primitive. Native frames show distinct
interaction states; overlays follow replayed movements. Gain inference and the
next action are schematic illustrations of the control process.}
\label{fig:overview}
\end{figure*}

Our evaluations reveal large performance gaps when successful control requires identifying hidden environment properties. To expose the evidence needed for adaptation, we organize the interface around \move: cursor movement, mouse-down, and mouse-up are separately executable, and each primitive returns an observation before the next agent decision (Figure~\ref{fig:overview}). We construct training trajectories that demonstrate iterative exploration and feedback-conditioned control, and train LatentLearner-9B on these trajectories. Ablations show that both action granularity and demonstration structure matter: overall success is 50.0\% with the full interface and data, compared with 19.6\% for bundled click/drag and 27.6\% when each movement phase contains a single supervised move.

Joint RL further improves control across the matched regimes (Section~\ref{sec:paired-rl}). Egocentric Drag and Ten-Choice success rises from 41.3\% and 40.7\% after SFT to 92.0\% and 75.3\%, while exocentric success remains 96.7\% and 100.0\%. Paired Rotation training reaches 27.3\% and 11.3\% on the inner and outer rings, improving on the inner-ring specialist and matching the outer-ring specialist. Finally, applying the same interaction format to standard grounding yields 67.37\% mean accuracy across five benchmarks after RL, compared with 66.72\% for conventional grounding RL (Section~\ref{sec:grounding-analysis}). Beyond success rates, the trained models exhibit feedback-conditioned multi-step exploration. In egocentric Drag, the Pearson correlation between error-normalized commands and ideal gain compensation $1/g$ increases from 0.610 after SFT to 0.803 after paired RL, indicating stronger compensation for episode-specific dynamics (Section~\ref{sec:paired-rl}). Together, these results connect learning from visual interaction to both adaptation within the benchmark and broader GUI grounding.

Our contributions are threefold:
\begin{itemize}
\item We introduce \textbf{LatentGUIWorld}, a benchmark for studying GUI control under hidden environment properties, exposing substantial failures of agents that cannot adapt from interaction feedback.

\item We propose a fine-grained action interface and feedback-rich training formulation that support iterative exploration and adaptation, and show through ablations that both action granularity and trajectory structure are critical.

\item We demonstrate that feedback-conditioned control can be learned jointly across interaction regimes and the same interaction format supports standard GUI grounding.

\end{itemize}

\section{Related Work}

\subsection{GUI Agent Benchmarks}
GUI agent benchmarks range from instruction-conditioned grounding, as in ScreenSpot-Pro~\cite{li2025screenspotpro}, to task execution in realistic environments, as in OSWorld and MobileBench-OL~\cite{xie2024osworld,wu2026mobilebenchol}. Recent efforts further extend task horizon: OSWorld 2.0 emphasizes long professional workflows involving cross-application coordination and sustained execution~\cite{yuan2026osworld2}. Meanwhile, CoAct-1 and OSWorld-MCP report benefits from combining GUI actions with programmatic tools~\cite{song2025coact1,jia2025osworldmcp}, and a matched GUI--CLI study finds that verifier-guided skill augmentation enables CLI agents to outperform their GUI counterparts~\cite{zhou2026guivscli}. These results invite reconsideration of how far increasing workflow length should drive GUI benchmark development when alternative interfaces can substantially simplify execution.

\begin{figure*}[t]
\centering
\resizebox{\linewidth}{!}{%
% Qualitative bands and category columns: no metric distances or task ranking.
\begin{tikzpicture}[x=1cm,y=1cm,
  font=\sffamily\fontsize{8}{10}\selectfont,
  labeltext/.style={align=center,inner sep=2pt},
  smallnote/.style={text=black!58,font=\sffamily\fontsize{7}{9}\selectfont},
  variant/.style={fill=white,rounded corners=2pt,inner xsep=7pt,
    inner ysep=4pt,align=center,font=\sffamily\fontsize{8}{10}\selectfont}]
  \definecolor{landscapeblue}{HTML}{2858A5}
  \definecolor{landscapedrag}{HTML}{287D8E}
  \definecolor{landscapechoice}{HTML}{A96A24}
  \definecolor{landscaperotation}{HTML}{8064A2}

  % The vertical axis has two qualitative bands, not a duration scale.
  \fill[black!2] (0,4.18) rectangle (14.8,5.77);
  \draw[black!18,line width=.5pt] (0,4.18) -- (14.8,4.18);
  \foreach \x in {5.92,10.18}
    \draw[black!16,densely dotted,line width=.5pt] (\x,0) -- (\x,5.77);
  \draw[->,black!75,line width=.75pt] (0,0) -- (15.02,0);
  \draw[->,black!75,line width=.75pt] (0,0) -- (0,6.00);
  \node[rotate=90,font=\sffamily\bfseries\fontsize{9}{11}\selectfont]
    at (-1.12,3.00) {Task horizon};
  \node[anchor=east,text=black!65] at (-.18,4.98) {Long};
  \node[anchor=east,text=black!65] at (-.18,2.06) {Short};

  % A grouped example list avoids implying benchmark-by-benchmark rankings.
  \node[labeltext,text=black!75,font=\sffamily\bfseries\fontsize{8}{10}\selectfont]
    at (3.72,5.50) {Known-interface workflows};
  \node[labeltext,text=black!65] at (3.72,4.83)
    {OSWorld / OSWorld 2.0\\VisualWebArena / AndroidWorld\\Windows Agent Arena};
  \node[labeltext,text=black!70] at (3.72,.63)
    {\textbf{Grounding / manipulation}\\
     ScreenSpot-Pro / UI-Vision\\ScreenDrag (ShowUI-$\pi$)};
  \begin{scope}[yshift=1.05cm]
  % One Short-band panel; its three rows denote families, not task durations.
  \filldraw[fill=landscapeblue!4,draw=landscapeblue!28,
    rounded corners=3pt,line width=.6pt]
    (.20,.24) rectangle (14.78,2.91);
  \node[anchor=west,text=landscapeblue,
    font=\sffamily\bfseries\fontsize{10}{12}\selectfont]
    at (.44,2.59) {LatentGUIWorld};
  \draw[black!40,line width=.65pt] (9.54,2.59) -- (10.06,2.59);
  \node[anchor=west,smallnote] at (10.16,2.59) {Matched variants (six tasks)};
  \draw[landscapeblue!15,line width=.4pt] (.44,2.28) -- (14.54,2.28);

  \node[anchor=west,text=landscapedrag,font=\sffamily\bfseries]
    at (.47,1.91) {Drag};
  \draw[landscapedrag!65,line width=.85pt] (3.72,1.91) -- (12.35,1.91);
  \node[variant,draw=landscapedrag!55] at (3.72,1.91) {Exocentric};
  \node[variant,draw=landscapedrag!55] at (12.35,1.91) {Egocentric};

  \node[anchor=west,text=landscapechoice,font=\sffamily\bfseries]
    at (.47,1.25) {Ten-Choice};
  \draw[landscapechoice!65,line width=.85pt] (8.05,1.25) -- (12.35,1.25);
  \node[variant,draw=landscapechoice!55] at (8.05,1.25) {Exocentric};
  \node[variant,draw=landscapechoice!55] at (12.35,1.25) {Egocentric};

  \node[anchor=west,text=landscaperotation,font=\sffamily\bfseries]
    at (.47,.59) {Rotation};
  \draw[landscaperotation!65,line width=.85pt] (11.3,.59) -- (13.39,.59);
  \node[variant,draw=landscaperotation!55,inner xsep=5pt]
    at (11.3,.59) {Inner-ring};
  \node[variant,draw=landscaperotation!55,inner xsep=5pt]
    at (13.39,.59) {Outer-ring};

  \end{scope}

  \foreach \x in {3.72,8.05,12.35}
    \draw[black!60,line width=.5pt] (\x,0) -- (\x,-.07);
  \node[anchor=north,labeltext] at (3.72,-.18)
    {\textbf{$I_0$}\\Prior knowledge +\\current context suffice};
  \node[anchor=north,labeltext] at (8.05,-.18)
    {\textbf{$I_1$}\\Probe outcome reveals\\hidden semantics};
  \node[anchor=north,labeltext] at (12.35,-.18)
    {\textbf{$I_2$}\\Action--outcome relation\\identifies dynamics};
  \node[font=\sffamily\bfseries\fontsize{9}{11}\selectfont]
    at (7.65,-1.40) {Minimum evidence for environment identification};
\end{tikzpicture}
}
\caption{Task horizon and the evidence required for environment identification are distinct dimensions. The six \lgw tasks form three matched pairs (connecting lines). Named benchmarks provide qualitative examples of selected known-interface workflows, grounding, or known-mapping manipulation~\cite{xie2024osworld,yuan2026osworld2,koh2024visualwebarena,rawles2024androidworld,bonatti2024windowsagentarena,li2025screenspotpro,nayak2025uivision,hu2025showuipi}. Placement characterizes those selected tasks; bands and within-band positions are schematic.}
\label{fig:benchmark-landscape}
\end{figure*}
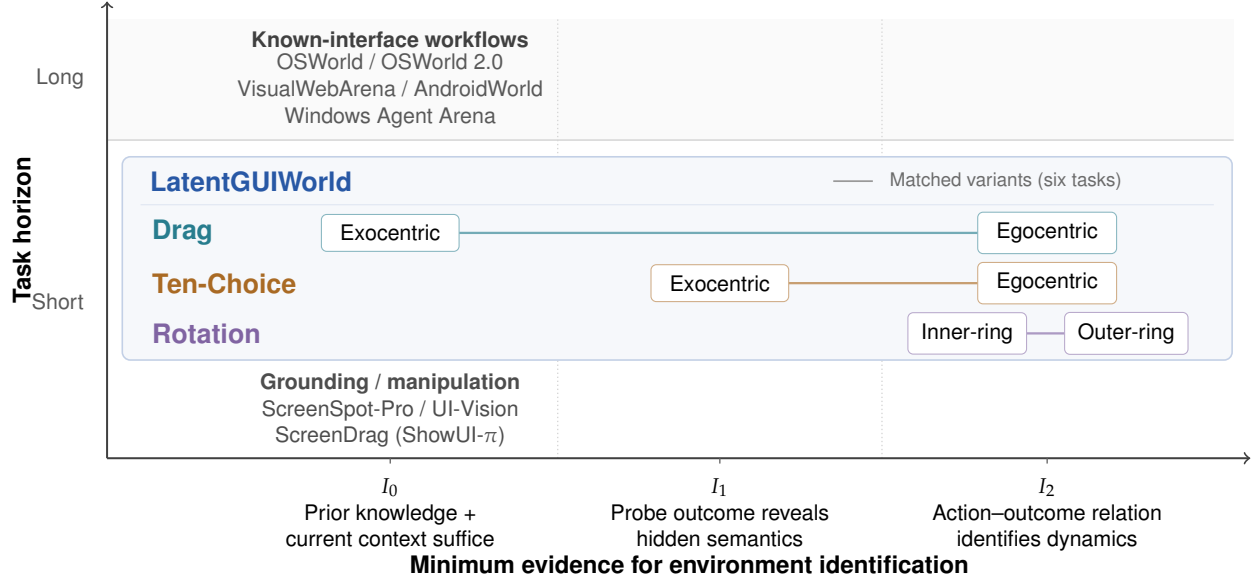

Other benchmarks examine challenges arising within interaction itself. ChartAct and CAP require agents to access information unavailable in a static view~\cite{huang2026chartact,xu2026cap}; DynamicGUIBench evaluates responses to interface changes~\cite{liu2026dynamicgui}; and Open CaptchaWorld and Next-Gen CAPTCHAs test perception and manipulation through interactive challenges~\cite{luo2025opencaptchaworld,liu2026nextgencaptcha}. GUI Knowledge Bench and EvoGUI further assess action effects and state-transition understanding from offline trajectories~\cite{shi2025guiknowledge,yang2026evogui}. Building on this direction, \lgw focuses on identifying hidden semantics and episode-specific control laws through an agent's own interactions. Its short, matched tasks make the evidence required for successful control an explicit evaluation dimension alongside task horizon. Figure~\ref{fig:benchmark-landscape} positions these interaction-evidence requirements alongside task horizon.

\subsection{Interaction and Adaptation in GUI Agents}
Existing GUI-agent action spaces support standalone cursor movement~\cite{qwen2025qwen3vl,lu2026qwencua}, and several methods use interaction feedback to refine decisions. GUI-Cursor turns grounding into iterative cursor movement, using visual feedback about cursor--target relations to refine subsequent movements~\cite{zhao2025guicursor}. ReCAP trains on self-correction trajectories derived from failures to support online error recovery in CAPTCHA solving~\cite{chen2026recap}.

Complementary work studies which interaction evidence to retain or acquire. DynamicUI selects informative frames from interaction videos to capture interface changes missed by isolated screenshots~\cite{liu2026dynamicgui}. ScreenSearch guides desktop exploration using an ambiguity signal based on differences in matched-action outcomes at similar screen states~\cite{solodko2026screensearch}. These approaches motivate our study of whether agents can use action-generated evidence to identify hidden semantics and episode-specific control laws, then use that evidence to guide subsequent control. Our controlled tasks separate direct semantic revelation from identification that requires interpreting an observation relative to the action that produced it.

To support this process, \move separates cursor movement from mouse-down and mouse-up events and returns a new observation after each primitive, allowing agents to acquire evidence and adjust control before committing through a click or release. Building on existing movement primitives, we combine this interface with aligned training trajectories to study how agents use movement to reveal hidden semantics, identify episode-specific dynamics, and guide subsequent decisions.

\section{\lgw}
\label{sec:benchmark}

\lgw studies episodic GUI environments in which the same task and pre-interaction context can conceal different environment configurations. The object to be identified is a task-relevant \emph{environment latent}: an episode-specific property that determines how the interface responds to an intervention and which subsequent control succeeds. We first formalize this object and classify tasks by the minimum evidence needed to identify it, then construct matched environments that instantiate the resulting taxonomy.

\subsection{Problem Definition and Taxonomy}
\label{sec:problem-definition}
Environment identification concerns a task-relevant, episode-specific latent $z$ that determines the downstream control required for success. The agent identifies its current value through interaction-generated evidence. Let
\[
H_t=(g,O_0,A_0,O_1,A_1,\ldots,O_{t-1},A_{t-1}),
\qquad
D_t=(H_t,O_t).
\]
Here $D_t$ is the complete observed context before the current intervention. All identifiability statements assume sufficient prior domain knowledge $K$, including GUI conventions, application rules, and the family of possible control laws. This knowledge is shared across episodes and excludes the realized hidden value $z$ of the current episode. The agent executes a probe action $A_t^{\mathrm{probe}}$ and receives
\[
O_{t+1}\sim T_z(D_t,A_t^{\mathrm{probe}}).
\]
$A_t^{\mathrm{probe}}$ is the agent's own action information naturally available to the agent because it generated the action; for a mouse movement, it includes the executed displacement, and may be written as $A_t^{\mathrm{probe}}=(\Delta x_t,\Delta y_t)$. Let $A_{t+1}^*(D_t,z)$ denote the next task-directed action required after the relevant evidence is acquired. We call $z$ an environment latent when it remains fixed within an episode, is not identifiable from $(K,D_t)$, changes this required downstream control, and can be identified from the consequences of the agent's own intervention. At a point of ambiguity, even sufficient prior knowledge cannot select the correct downstream control from the observed context. Formally, under the same $K$, two latent values can satisfy
\[
\begin{aligned}
z_1&\neq z_2, & D_t(z_1)&=D_t(z_2),\\
T_{z_1}(D_t,A_t^{\mathrm{probe}})&\neq T_{z_2}(D_t,A_t^{\mathrm{probe}}), &
A_{t+1}^*(D_t,z_1)&\neq A_{t+1}^*(D_t,z_2),
\end{aligned}
\]
Table~\ref{tab:interaction-levels} defines three levels according to the minimum evidence sufficient to resolve this ambiguity.

\begin{table}[t]
\centering
\scriptsize
\setlength{\tabcolsep}{3.2pt}
\renewcommand{\arraystretch}{1.18}
\begin{tabularx}{\linewidth}{@{}c >{\raggedright\arraybackslash}p{0.17\linewidth} >{\raggedright\arraybackslash}p{0.35\linewidth} >{\raggedright\arraybackslash}X@{}}
\toprule
\textbf{Level} & \textbf{Before intervention} & \textbf{Minimum identifying evidence} & \textbf{Dependency} \\
\midrule
$\mathbf{I_0}$ &
$K$ and $D_t$ suffice. &
Prior knowledge and observed context support control; multiple execution steps and screen updates are allowed. &
$\boxed{D_t=(H_t,O_t)\rightarrow A_t^*}$ \\
\addlinespace
$\mathbf{I_1}$ &
$K$ and $D_t$ do not suffice. &
The probe outcome directly reveals a semantic latent: use \emph{what the action revealed}. &
$\boxed{D_t\xrightarrow{A_t^{\mathrm{probe}}}O_{t+1}\rightarrow z_{\mathrm{sem}}\rightarrow A_{t+1}^*}$ \\
\addlinespace
$\mathbf{I_2}$ &
$K$, $D_t$, and a probe outcome alone do not identify the dynamics. &
The self-action and induced transition jointly identify a dynamics latent: use \emph{what the action caused}. &
$\boxed{(O_t,A_t^{\mathrm{probe}},O_{t+1})\rightarrow z_{\mathrm{dyn}}\rightarrow A_{t+1}^*}$ \\
\bottomrule
\end{tabularx}
\caption{Evidence taxonomy for environment identification. $D_t=(H_t,O_t)$ is the observed pre-intervention context, $A_t^{\mathrm{probe}}$ is the self-executed probe, and $A_{t+1}^*$ is the subsequent task-directed action. Sufficiency is evaluated with prior domain knowledge $K$, excluding the current episode's hidden latent. A task is assigned the lowest level sufficient for reliable success under a fixed benchmark contract.}
\label{tab:interaction-levels}
\end{table}

Two distinctions are important. In $I_1$, the probe result directly reveals the missing semantic value, so the action mainly determines \emph{what is revealed}. In $I_2$, the observation must be interpreted relative to the self-executed action, so the agent reasons about \emph{what its action caused}; we call this capability \emph{online interaction dynamics identification}. Because the executed action parameters are part of the agent's own history, $I_2$ conditions on the complete intervention--response record.

Let $\Pi_{\epsilon,B}(\mathcal{T})$ be the policies with access to $K$ that solve task $\mathcal{T}$ with error at most $\epsilon$ under interaction budget $B$. We assign a task the lowest evidence level sufficient for reliable success:
\begin{equation}
I(\mathcal{T})=\min_{\pi\in\Pi_{\epsilon,B}(\mathcal{T})}\ \max_{t\in\tau_\pi} I_t.
\end{equation}
This minimum assigns a task to $I_0$ when prior knowledge and observed context suffice, even if a less knowledgeable agent explores unnecessarily. An $I_0$ task may contain many actions and post-action observations; its execution requires no identification of a hidden episode-specific property. We evaluate control through task success within a fixed interaction budget, scoring the outcome independently of verbal estimates of the latent. Paired trajectories in Appendix~\ref{app:paired-policy-cases} illustrate how successful control uses the different feedback generated from identical initial observations.

The level concerns identifying the goal-directed choice or control before the missing evidence is acquired. Knowing which probe to execute is insufficient: in Ten-Choice, an agent may know that it should hover while remaining unable to identify the candidate it must ultimately select. Once the latent is identified and stored in $H_t$, later decisions may revert to $I_0$. Evidence level is independent of history length: memory, planning, cross-application navigation, and completed subgoals may all be contained in $H_t$ while a particular decision remains $I_0$.

An $I_1$ construction pairs pre-interaction aliased instances whose required action becomes identifiable from the probe outcome. Heterogeneous suites may contain decisions at multiple levels, so we assign levels to explicit task constructions under a fixed contract.

\subsection{Benchmark Construction}
\begin{figure*}[p]
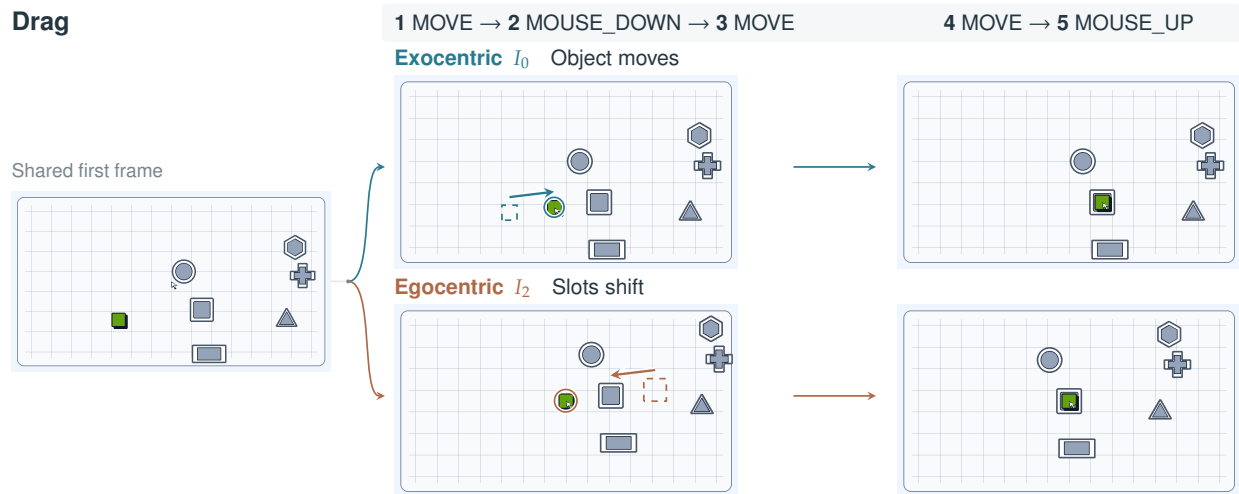

% A page-height panel layout absorbs spare space between task families instead
% of leaving it below the figure. The original standalone artwork is preserved.
\begin{minipage}[t][\dimexpr\textheight-2pt\relax][s]{\textwidth}
\centering
\begingroup
\def\BenchmarkPairPanelIndex{1}
\def\BenchmarkPairPanelBottom{17.35}\def\BenchmarkPairPanelTop{25.65}
\input{figures/benchmark-pair-grouped-artwork.tex}\par
\endgroup
\vfill
\begingroup
\def\BenchmarkPairPanelIndex{2}
\def\BenchmarkPairPanelBottom{8.80}\def\BenchmarkPairPanelTop{17.10}
\input{figures/benchmark-pair-grouped-artwork.tex}\par
\endgroup
\vfill
\begingroup
\def\BenchmarkPairPanelIndex{3}
\def\BenchmarkPairPanelBottom{0.25}\def\BenchmarkPairPanelTop{8.55}
\input{figures/benchmark-pair-grouped-artwork.tex}\par
\endgroup
\caption{\textbf{Shared first frames, different interaction mechanisms.}
Action headers apply to both variants; coordinates may differ. Each numbered
primitive executes separately, with intermediate frames omitted. Middle and right
panels follow the actions above them. Arrows mark motion; dashed marks indicate
pre-MOVE positions (Ten-Choice: action 1; Drag: action 3) or Rotation's fixed inner
boundary. Scripted replay trajectories illustrate each control mechanism.
Drag uses a matched-reset pair.}
\label{fig:task-grid}
\end{minipage}
\end{figure*}

\lgw contains three interactive environments---Drag, Ten-Choice, and Rotation---with two matched variants in each environment. Each pair preserves its high-level task objective while varying the interaction mechanism or control law. In Drag and Ten-Choice, the exocentric variant uses a movable cursor over a fixed scene, whereas the egocentric variant keeps the interaction point at the screen center and moves the viewpoint relative to the scene. Figure~\ref{fig:task-grid} illustrates the environments and successful terminal states.

\subsubsection*{Drag}
The agent must drag a solid object into its matching outline among several candidate slots.

\paragraph{\textbf{Variant A:} Exocentric control ($I_0$).}
A free cursor moves a visible object toward its matching slot under a fixed camera and a globally known one-to-one screen-space mapping. There is no task-relevant environment latent to identify: the visible object and slot locations already determine the next held-object movement, so the task is $I_0$.

\Needspace{4\baselineskip}
\paragraph{\textbf{Variant B:} Egocentric control ($I_2$).}
The matched variant uses a center-reticle, viewpoint-relative control law. Its dynamics environment latent is the two-axis held-object transfer gain $z_{\mathrm{dragEgo}}=(g_x,g_y)$: for an executed cursor displacement $\Delta u_t$, the held object moves by $p_{t+1}-p_t=\operatorname{diag}(g_x,g_y)\Delta u_t$.
The evaluated variant uses equal positive gains on both axes, whose magnitude varies across episodes and is not shown in the initial frame. The same egocentric movement can therefore transport the held object by different amounts in otherwise identical samples. The agent must identify $z_{\mathrm{dragEgo}}$ by relating its executed displacement to the resulting visual transition, $(O_t,\Delta u_t,O_{t+1})$, before it can place the object efficiently, making the variant $I_2$.

Each Drag variant contains 75 IID cases using training shapes and 75 OOD cases using held-out shapes.

\subsubsection*{Ten-Choice}
\paragraph{\textbf{Variant A:} Exocentric reveal ($I_1$).}
Ten visually identical candidates hide messages from $\mathcal{M}=\{m_1,\ldots,m_{10}\}$. The semantic environment latent is an episode-specific permutation $z_{\mathrm{10}}^{\mathrm{sem}}=\pi$ that maps candidate indices to messages. The instruction specifies a target message $m^*$. Before interaction, neither the initial frame nor the history identifies which candidate contains $m^*$. Moving onto candidate $i$ renders $\pi(i)$ directly in $O_{t+1}$. Because the probe result itself reveals the task-relevant value without requiring action--effect comparison, this variant is $I_1$; after finding $\pi(i)=m^*$, the final click is an $I_0$ decision.

\paragraph{\textbf{Variant B:} Egocentric reveal ($I_2$).}
The matched variant retains the hidden permutation $\pi$ and adds an episode-specific egocentric viewpoint law. Its dynamics environment latent is the gain $z_{\mathrm{10Ego}}^{\mathrm{dyn}}=(s_x,s_y)$, which maps movement $\Delta u_t$ to viewpoint change $v_{t+1}-v_t=\operatorname{diag}(s_x,s_y)\Delta u_t$.
The complete environment latent is $(\pi,z_{\mathrm{10Ego}}^{\mathrm{dyn}})$. The permutation still requires $I_1$ revelation, but the task's required level is $I_2$: the agent must compare its movement with the displacement of the candidate field to infer $(s_x,s_y)$ before it can steer efficiently to the candidates and reveal their messages.

\Needspace{4\baselineskip}
\subsubsection*{Rotation}
The agent must rotate the controlled image region into alignment with the surrounding image.

\paragraph{\textbf{Variant A:} Inner-ring control ($I_2$).}
The inner-ring variant uses an episode-specific signed scalar gain $z_{\mathrm{rot}}=k$. A horizontal movement $\Delta x_t$ produces object rotation $\Delta\theta_t=k\Delta x_t$, so the same initial frame can require different movement magnitudes or directions. From one probe, the agent can estimate $k$ by comparing the observed rotation change with its executed displacement and use that estimate to calibrate the next correction. Although $O_{t+1}$ alone can verify whether the object is under- or over-rotated, identifying $z_{\mathrm{rot}}$ requires the full transition $(O_t,\Delta x_t,O_{t+1})$, making the task $I_2$.

\paragraph{\textbf{Variant B:} Outer-ring control ($I_2$).}
The matched variant moves the control to an outer ring. Its dynamics environment latent is the signed angular coupling $z_{\mathrm{outer}}=k_{\mathrm{outer}}$, which maps the ring displacement $\Delta\phi_t$ to object rotation $\Delta\theta_t=k_{\mathrm{outer}}\Delta\phi_t$.
The magnitude of $k_{\mathrm{outer}}$ determines response sensitivity, while its sign determines whether the object follows or opposes the outer-ring motion. The agent must compare the applied ring displacement with the observed object rotation to infer $z_{\mathrm{outer}}$; a post-move image without the self-action is sufficient for verification but not calibrated control.

\paragraph{Episode construction and test splits.}
The fixed test suite contains 150 episodes per variant, totaling 900 episodes. A separate validation set contains the same number of episodes per variant (900 in total) and is used for checkpoint selection. Test cases vary in scene configuration and task-specific parameters. Semantic assignments and control parameters are assigned during dataset construction and remain fixed throughout each trajectory; task-relevant latents are not disclosed in the initial observation. Ten-Choice uses message content disjoint from its task-training data, and Rotation uses disjoint background images. Variant pairs share scene content and task objectives, although some egocentric Drag cases place the target initially offscreen and therefore do not share identical first frames. Detailed sampling configurations, train--test separation, and pairing statistics are provided in Appendix~\ref{app:benchmark-construction}.

\section{Learning from \move}

\subsection{\move-Centric Action Schema}
We organize the action schema around $\operatorname{MOVE}(p)$, which moves the cursor to a specified position, together with coordinate-free $\operatorname{CLICK}$, $\operatorname{MOUSE\_DOWN}$, and $\operatorname{MOUSE\_UP}$ events that act at the current cursor position. After each independently executed action, the environment returns a new observation before the model predicts the next action. Coordinate-bearing $\operatorname{CLICK}(x,y)$ and $\operatorname{DRAG}(p,q)$ macros remain available, but their internal movements do not introduce agent decision points. When expressed through \move primitives, click and drag interactions expose these intermediate observations:
\begin{equation}
\operatorname{CLICK}(p)\;\rightsquigarrow\;
[\operatorname{MOVE}\rightarrow O]^+\rightarrow\operatorname{CLICK}\rightarrow O.
\end{equation}
\begin{equation}
\operatorname{DRAG}(p,q)\;\rightsquigarrow\;
[\operatorname{MOVE}\rightarrow O]^+\rightarrow\operatorname{MOUSE\_DOWN}\rightarrow O
\rightarrow[\operatorname{MOVE}\rightarrow O]^+\rightarrow\operatorname{MOUSE\_UP}\rightarrow O.
\end{equation}

Here each $O$ denotes a fresh observation, and $[\operatorname{MOVE}\rightarrow O]^+$ denotes one or more individually decided movements, each followed by its observation. For clicking, movement ends at $p$; for dragging, contact begins at $p$ and release occurs at $q$. At each boundary, the model conditions its next action on the observed transition and may continue moving, gather further evidence, or commit through a click or release. Each movement provides an action--observation transition $(O_t,A_t^{\mathrm{move}},O_{t+1})$ for the next decision. This supports direct use of revealed semantics in $I_1$ and action--effect comparison for dynamics identification in $I_2$. We refer to this feedback-conditioned control process as \emph{learning from \move}.

\subsection{Training Data Construction}
\label{sec:move-data}
We construct multi-step \move demonstrations across all six \lgw variants using rule-based scripts. These scripts execute primitive actions and record the resulting screenshot after each action. For interaction-dependent tasks, they inspect revealed content or estimate movement responses from observed transitions, then use this feedback to choose subsequent moves before clicking or releasing. We retain successful trajectories as alternating sequences of observations and actions, supervising exploration and correction during free or held movement.

A teacher model adds a short Think rationale for each prescribed action from its pre-action observations, preceding responses, and the action itself, without changing the sequence. The resulting demonstrations combine rule-generated action sequences with teacher-generated rationales.

Multi-step \move data can also be obtained from existing GUI and computer-use demonstrations by decomposing actions and supplying intermediate observations through environment replay or intermediate-frame construction, where feasible. We explore this extension using grounding data in Section~\ref{sec:grounding-analysis}.

\subsection{Model Training}
\label{sec:model-training}
We first train the policy through supervised fine-tuning on demonstration trajectories. Following Section~\ref{sec:problem-definition}, let $D_t=(H_t,O_t)$ denote the observed context before action $A_t$. Each training example pairs this context with the demonstrated response $y_t$, and we minimize the autoregressive objective:
\begin{equation}
\mathcal{L}_{\mathrm{SFT}}(\theta)
=-\mathbb{E}_{(D_t,y_t)\sim\mathcal{D}}
\log\pi_\theta(y_t\mid D_t).
\end{equation}
For \lgw, $y_t$ contains a Think rationale followed by the primitive action; the grounding extension uses action-only targets. This objective supervises the next response using observations and responses available up to the current decision.

We further explore reinforcement learning initialized from the SFT policy. The policy samples trajectories through online interaction, receiving a new observation after each action and conditioning subsequent decisions on the updated context. We use Group Relative Policy Optimization (GRPO) to optimize trajectory rewards based on task success and action-format compliance. Experiment-specific setups are described in Sections~\ref{sec:experimental-setup} and~\ref{sec:grounding-analysis}; training details and hyperparameters are provided in Appendix~\ref{app:training-details}.

\section{Experiments}

\subsection{Experimental Setup}
\label{sec:experimental-setup}
\paragraph{Tasks and data.}
We evaluate the six environments in Section~\ref{sec:benchmark}, organized into matched Drag, Ten-Choice, and Rotation pairs. The test suite contains 150 episodes per environment. LatentLearner-9B is trained jointly on 24,000 successful demonstration trajectories, with 4,000 per environment, yielding 154,713 supervised examples. Data separation and training details are given in Appendices~\ref{app:benchmark-construction} and~\ref{app:benchmark-training}.

\paragraph{Evaluation interfaces.}
GPT-6 Astra follows the officially recommended code-execution interface, and Claude Opus 5 uses the official computer-use schema. GPT-5.6-Sol and Kimi-K3 use their official schemas with the tools needed for task completion. DeepSeek-V4.1-Flash uses our six-action JSON schema, as no official computer-use schema was available. Appendix~\ref{app:frontier-protocols} specifies the tools and evaluated variants. For GUI-specialized models, we follow the official evaluation schemas and provide additional tools when the official action space is insufficient to complete the tasks.

\paragraph{Models and training.}
We compare GUI agents based on general-purpose models with GUI-specialized agents, as grouped in Table~\ref{tab:main}. LatentLearner-9B uses Qwen3.5-9B~\cite{qwen2026qwen35} with joint supervised fine-tuning in the \move format. Each training example conditions the next action on the task instruction, up to the latest three screenshots, and preceding assistant actions. The reported benchmark model generates with-think responses and receives no RL updates. One checkpoint is used across all six environments, selected by overall success rate on the separate validation set. Training hyperparameters and checkpoint-selection details are given in Appendix~\ref{app:benchmark-training}.

Appendix~\ref{app:main-result-prompt} gives the task prompt and multimodal history used by our SFT models and the raw Qwen baseline.

\paragraph{Metrics.}
We report episode success for each environment, the macro-average over the four $I_2$ tasks, and the macro-average over all six tasks. Rotation uses strict first-release scoring: a failed first release cannot be recovered by a later release or by final-state success at the action limit. Recorded endpoint, protocol, and executor failures count as nonsuccesses. Drag and Ten-Choice pairs compare different evidence requirements, whereas both Rotation variants are $I_2$ and compare control geometries.

\subsection{Benchmark Performance and Control Behavior}
\label{sec:benchmark-results}
We evaluate whether strong performance on observed-state and semantic-reveal tasks extends to their matched dynamics-dependent variants.

\begin{table*}[!t]
\centering
\caption{Success rate (\%) across six environments in three matched pairs. $\Delta_{\mathrm{pair}}$ is the second variant minus the first in each pair, in percentage points, computed before rounding; both Rotation variants are $I_2$. $I_2$ Avg. averages the four $I_2$ tasks; Overall averages all six. Rotation uses strict first-release scoring. Each reported row uses one model checkpoint across environments. SFT checkpoints are selected on the separate validation set; all scores shown are test results. Bold and underline mark the best and second-best results per column, excluding pairwise differences.}
\label{tab:main}
\footnotesize
\setlength{\tabcolsep}{2pt}
\renewcommand{\arraystretch}{1.20}
\begin{tabularx}{\textwidth}{Xccccccccccc}
\toprule
\multirow{2}{*}{\textbf{GUI agent}} & \multicolumn{3}{c}{\textbf{Drag}} & \multicolumn{3}{c}{\textbf{Ten-Choice}} & \multicolumn{3}{c}{\textbf{Rotation}} & \multirow{2}{*}{\makecell{\textcolor{mosiblue}{\textbf{$I_2$}}\\\textcolor{mosiblue}{\textbf{Avg.}}}} & \multirow{2}{*}{\textcolor{mosiblue}{\textbf{Overall}}}\\
\cmidrule(lr){2-4}\cmidrule(lr){5-7}\cmidrule(lr){8-10}
 & \makecell{Exocentric\\($I_0$)} & \makecell{Egocentric\\($I_2$)} & \cellcolor{mosiblue!7}$\Delta_{\mathrm{pair}}$ & \makecell{Exocentric\\($I_1$)} & \makecell{Egocentric\\($I_2$)} & \cellcolor{mosiblue!7}$\Delta_{\mathrm{pair}}$ & \makecell{Inner-ring\\($I_2$)} & \makecell{Outer-ring\\($I_2$)} & \cellcolor{mosiblue!7}$\Delta_{\mathrm{pair}}$ & & \\
\midrule
\rowcolor{black!4}
\multicolumn{12}{l}{\strut\textit{GUI agents based on general-purpose models}} \\
GPT-5.6-Sol~\cite{openai2026gpt56sol} & \textbf{100.0} & 4.7 & \cellcolor{mosiblue!7}$-95.3$ & \textbf{100.0} & 0.7 & \cellcolor{mosiblue!7}$-99.3$ & 18.0 & 5.3 & \cellcolor{mosiblue!7}$-12.7$ & 7.2 & 38.1 \\
GPT-6 Astra~\cite{openai2026gpt6astra} & \textbf{100.0} & 2.0 & \cellcolor{mosiblue!7}$-98.0$ & \textbf{100.0} & \textbf{56.0} & \cellcolor{mosiblue!7}$-44.0$ & \textbf{65.3} & \textbf{56.7} & \cellcolor{mosiblue!7}$-8.7$ & \textbf{45.0} & \textbf{63.3} \\
Claude Opus 5~\cite{anthropic2026opus5} & \textbf{100.0} & 0.0 & \cellcolor{mosiblue!7}$-100.0$ & \textbf{100.0} & 1.3 & \cellcolor{mosiblue!7}$-98.7$ & \underline{32.7} & \underline{16.7} & \cellcolor{mosiblue!7}$-16.0$ & 12.7 & 41.8 \\
Kimi-K3~\cite{moonshot2026kimik3} & \textbf{100.0} & \underline{6.0} & \cellcolor{mosiblue!7}$-94.0$ & \textbf{100.0} & 3.3 & \cellcolor{mosiblue!7}$-96.7$ & 25.3 & 6.0 & \cellcolor{mosiblue!7}$-19.3$ & 10.2 & 40.1 \\
DeepSeek-V4.1-Flash~\cite{deepseek2026v41flash} & 80.0 & 0.0 & \cellcolor{mosiblue!7}$-80.0$ & 19.3 & 1.3 & \cellcolor{mosiblue!7}$-18.0$ & 0.7 & 0.7 & \cellcolor{mosiblue!7}$0.0$ & 0.7 & 17.0 \\
\addlinespace[3pt]
\rowcolor{black!4}
\multicolumn{12}{l}{\strut\textit{GUI-specialized agents}} \\
Qwen-CUA~\cite{lu2026qwencua} & \underline{99.3} & 0.7 & \cellcolor{mosiblue!7}$-98.7$ & 92.7 & 1.3 & \cellcolor{mosiblue!7}$-91.3$ & 0.0 & 0.7 & \cellcolor{mosiblue!7}$0.7$ & 0.7 & 32.4 \\
GUI-Owl-1.5-32B-Think~\cite{xu2026mobileagent35} & 59.3 & 0.0 & \cellcolor{mosiblue!7}$-59.3$ & 8.0 & 1.3 & \cellcolor{mosiblue!7}$-6.7$ & 1.3 & 1.3 & \cellcolor{mosiblue!7}$0.0$ & 1.0 & 11.9 \\
UI-TARS-1.5-7B~\cite{qin2025uitars} & 34.0 & 0.0 & \cellcolor{mosiblue!7}$-34.0$ & 8.7 & 1.3 & \cellcolor{mosiblue!7}$-7.3$ & 2.7 & 2.7 & \cellcolor{mosiblue!7}$0.0$ & 1.7 & 8.2 \\
DART-GUI-7B~\cite{li2025dartgui} & 21.3 & 0.0 & \cellcolor{mosiblue!7}$-21.3$ & 9.3 & 0.7 & \cellcolor{mosiblue!7}$-8.7$ & 3.3 & 3.3 & \cellcolor{mosiblue!7}$0.0$ & 1.8 & 6.3 \\
EvoCUA-32B~\cite{xue2026evocua} & 7.3 & 0.0 & \cellcolor{mosiblue!7}$-7.3$ & 14.7 & 0.0 & \cellcolor{mosiblue!7}$-14.7$ & 2.0 & 2.0 & \cellcolor{mosiblue!7}$0.0$ & 1.0 & 4.3 \\
OpenCUA-72B~\cite{wang2025opencua} & 32.7 & 0.7 & \cellcolor{mosiblue!7}$-32.0$ & 9.3 & 0.7 & \cellcolor{mosiblue!7}$-8.7$ & 0.0 & 0.0 & \cellcolor{mosiblue!7}$0.0$ & 0.3 & 7.2 \\
\addlinespace[3pt]
\rowcolor{black!4}
\multicolumn{12}{l}{\strut\textit{Backbone baseline and our models}} \\
Qwen3.5-9B (raw)$^{\dagger}$ & 0.7 & 0.0 & \cellcolor{mosiblue!7}$-0.7$ & 0.0 & 0.0 & \cellcolor{mosiblue!7}$0.0$ & 0.0 & 0.0 & \cellcolor{mosiblue!7}$0.0$ & 0.0 & 0.1 \\
\rowcolor{mosiblue!7}
\textbf{LatentLearner-9B} & 96.0 & \textbf{41.3} & \cellcolor{mosiblue!7}$-54.7$ & \underline{99.3} & \underline{40.7} & \cellcolor{mosiblue!7}$-58.7$ & 11.3 & 11.3 & \cellcolor{mosiblue!7}$0.0$ & \underline{26.2} & \underline{50.0} \\
\bottomrule
\end{tabularx}
\par\smallskip
\begin{minipage}{\textwidth}\scriptsize
$^{\dagger}$Public weights without project SFT/RL, evaluated with our with-think training prompt (128 output tokens/turn).
\end{minipage}
\end{table*}

\FloatBarrier
Table~\ref{tab:main} shows that strong performance on $I_0/I_1$ tasks does not ensure success on their matched $I_2$ variants. GPT-6 Astra achieves 100\% on exocentric Drag and Ten-Choice, but only 2.0\% and 56.0\% on their egocentric counterparts, despite having the highest $I_2$ average. GUI-specialized agents exhibit similar gaps. Several achieve only 8.0--14.7\% on exocentric Ten-Choice ($I_1$), indicating difficulty acquiring the hidden target semantics. Qwen-CUA is an exception at 92.7\%, yet falls to 1.3\% on the paired $I_2$ variant; its Drag success likewise drops from 99.3\% ($I_0$) to 0.7\% ($I_2$). The matched comparisons expose distinct difficulties in acquiring hidden semantics and adapting movement to hidden dynamics, even when the task objective is unchanged.

LatentLearner-9B achieves 50.0\% overall success and 26.2\% on $I_2$, with 96.0\% and 99.3\% on the $I_0/I_1$ variants and 41.3\% and 40.7\% on egocentric Drag and Ten-Choice. Rotation remains difficult at 11.3\% on each variant. In egocentric Drag, Pearson correlation between error-normalized commands and ideal gain compensation $1/g$ is $r=0.610$, indicating a tendency to compensate for gain. Appendix~\ref{app:mixed-sft-behavior} reports behavioral measurements and coverage.

\FloatBarrier
\subsection{Frontier-Agent Failure Analysis}
\label{sec:frontier-failures}
\paragraph{Acting before observing feedback.}
In the inspected trajectories, agents sometimes commit before movement-induced feedback informs their next decision. In Rotation, a bundled drag releases before the model observes its effect. In egocentric Drag, agents may move and immediately press without checking how the scene has shifted relative to the fixed interaction center. Such sequences can suffice under known mappings, but leave no opportunity to adapt the next action when the movement response is uncertain.

\paragraph{Feedback-based correction remains unreliable.}
Sol, Astra, Opus, and Kimi also exhibit feedback-conditioned control: maintaining the press, observing the resulting rotation, adjusting subsequent movements, and then releasing. These behaviors show that agents can use intermediate feedback, but their corrections and stopping decisions remain unreliable. Some Astra trajectories reach an aligned state yet continue moving before release, turning a potentially successful attempt into failure. Effective control requires coordinating observation, adjustment, and commitment.

\paragraph{Sensitivity to control geometry.}
Sol, Astra, Opus, and Kimi all perform worse on outer-ring than inner-ring Rotation (Table~\ref{tab:main}), although both variants are $I_2$ and share the alignment objective. Adaptation remains sensitive to control geometry.

\section{Analysis}
We first examine whether explicit exploration and perspective instructions improve a frontier agent without training. We then study three aspects of learning from \move: joint RL across interaction regimes, the role of action granularity and intermediate feedback, and compatibility with standard GUI grounding. These analyses connect task success to the evidence available to the model and the behavior learned from that evidence.

\subsection{Training-Free Prompt Ablation}
\label{sec:prompt-ablation}
We append ``Before attempting to solve the task, first explore the environment.'' and ``The current environment is egocentric.'' to GPT-6 Astra's baseline task instruction (Appendix~\ref{app:main-result-prompt}). We evaluate the first 50 episode IDs from each egocentric environment in the frozen test set and compare baseline and prompted performance on these same episodes. The model, native tool interface, task instances, initial observations, and other settings remain fixed.

\begin{table}[!ht]
\centering
\small
\begin{tabular}{lccc}
\toprule
\textbf{Egocentric task} & \textbf{Baseline} & \textbf{Prompted} & \textbf{Change} \\
\midrule
Ten-Choice & 28/50 (56\%) & 41/50 (82\%) & +26 pp \\
Drag & 1/50 (2\%) & 11/50 (22\%) & +20 pp \\
\bottomrule
\end{tabular}
\caption{Training-free prompt ablation on the same 50 episodes per environment. Prompted adds both exploration and egocentric instructions; pp denotes percentage points. These subset results are reported separately from the full evaluation in Table~\ref{tab:main}.}
\label{tab:prompt-ablation}
\end{table}

Table~\ref{tab:prompt-ablation} shows gains of 26 percentage points on Ten-Choice and 20 on Drag without additional training. These gains reveal a gap between autonomous and explicitly guided adaptation under the same interaction interface. The added cues provide task-specific guidance beyond tool-use instructions; Drag nevertheless reaches only 22\% success. These results show that even a frontier model struggles to identify and adapt to environment dynamics, despite explicit guidance.

\subsection{Joint RL Across Interaction Regimes}
\label{sec:paired-rl}
Can joint RL improve control across a matched pair, and how does its performance compare with environment-specific RL? We compare six single-environment RL specialists of LatentLearner-9B with three paired-RL models, one for each task family. Each specialist receives RL updates on one variant, whereas each paired model receives updates on both. We evaluate success on the same 150 held-out instances per variant, using greedy decoding, a 12-turn budget, and strict first-release scoring. Each paired model is evaluated on both variants with a single checkpoint. Appendix~\ref{app:paired-rl-selection} gives training settings and model selection.

\begin{table}[!ht]
\centering
\small
\setlength{\tabcolsep}{5pt}
\begin{tabular}{llrrr}
\toprule
\textbf{Pair} & \textbf{Test variant} & \textbf{Single-env. RL} & \textbf{Paired RL} & $\boldsymbol{\Delta}$ \\
\midrule
Drag & Exocentric & 96.67 & 96.67 & 0.00 \\
 & Egocentric & 93.33 & 92.00 & $-1.33$ \\
\midrule
Ten-Choice & Exocentric & 100.00 & 100.00 & 0.00 \\
 & Egocentric & 78.00 & 75.33 & $-2.67$ \\
\midrule
Rotation & Inner-ring & 24.00 & \textbf{27.33} & $+3.33$ \\
 & Outer-ring & 11.33 & 11.33 & 0.00 \\
\bottomrule
\end{tabular}
\caption{Single-environment versus paired RL initialized from LatentLearner-9B: success rates (\%). Each single-environment entry uses a separate specialist; the two paired-RL entries within each family use the same checkpoint. $\Delta$ is paired minus single-environment success in percentage points, computed before rounding.}
\label{tab:within-pair}
\end{table}

Table~\ref{tab:within-pair} shows that joint RL improves over single-environment RL on inner-ring Rotation and matches it on outer-ring Rotation. Drag and Ten-Choice instead pair an $I_0$ or $I_1$ exocentric task with its $I_2$ egocentric variant. Paired RL matches the $I_0/I_1$ specialists and remains close to, but slightly below, the $I_2$ specialists, while substantially improving $I_2$ success over mixed SFT (Table~\ref{tab:main}). These results support joint competence across the control regimes within each task family.

The behavioral analysis also shows stronger gain-dependent command scaling after paired RL. In egocentric Drag, Pearson correlation between error-normalized commanded displacement and the ideal gain-compensation coefficient $1/g$ increases from $r=0.610$ for mixed SFT to $r=0.803$ for paired RL. Success on the same evaluation suite rises from 41.3\% to 92.0\%. For a given target error, gain compensation requires larger commands under low gain and smaller commands under high gain. The increased correlation indicates closer alignment with this control requirement. Together, higher success and stronger gain compensation support more effective adaptation to episode-specific control dynamics. Appendix~\ref{app:mixed-sft-behavior} reports movement coverage, filtered results, and the comparison on shared episode IDs.

Two recorded test pairs further illustrate feedback-conditioned control. With identical initial screenshots, task prompts, and first actions, the policy receives different feedback and changes its subsequent actions under the two interaction mechanisms. Appendix~\ref{app:paired-policy-cases} presents the complete Ten-Choice and Drag cases.

\FloatBarrier

\subsection{Interaction Paradigm and Training Data Ablations}
\label{sec:interaction-ablation}
Table~\ref{tab:schema-ablation} compares LatentLearner-9B with ablations of its interaction paradigm and training data. All variants use Qwen3.5-9B with joint six-environment SFT and with-think responses.

\begin{table}[!ht]
\centering
\begin{threeparttable}
\caption{Success rates (\%) for interaction-paradigm and training-data ablations. Overall averages all six environments. Evaluation uses a 12-turn budget and strict first-release scoring. Bold and underlined values denote the best and second-best results per column.}
\label{tab:schema-ablation}
\footnotesize
\setlength{\tabcolsep}{2pt}
\begin{tabular*}{\textwidth}{@{\extracolsep{\fill}}lccrrrrrrr@{}}
\toprule
\multirow{2}{*}{\textbf{Setting}} & \multirow{2}{*}{\makecell{\textbf{\move}\\\textbf{interface}}} & \multirow{2}{*}{\makecell{\textbf{Iterative}\\\textbf{interaction data}}} & \multicolumn{2}{c}{\textbf{Drag}} & \multicolumn{2}{c}{\textbf{Ten-Choice}} & \multicolumn{2}{c}{\textbf{Rotation}} & \multirow{2}{*}{\textbf{Overall}} \\
\cmidrule(lr){4-5}\cmidrule(lr){6-7}\cmidrule(lr){8-9}
 & & & Exocentric & Egocentric & Exocentric & Egocentric & Inner-ring & Outer-ring & \\
\midrule
Paradigm ablation & $\times$ & $\times$ & \textbf{100.0} & 0.0 & 2.0 & 2.0 & 7.3 & \underline{6.0} & 19.6 \\
\addlinespace[3pt]
Data ablation & $\checkmark$ & $\times$ & 95.3 & \underline{12.0} & \underline{40.7} & \underline{3.3} & \underline{8.0} & \underline{6.0} & \underline{27.6} \\
\addlinespace[3pt]
\textbf{LatentLearner-9B} & $\checkmark$ & $\checkmark$ & \underline{96.0} & \textbf{41.3} & \textbf{99.3} & \textbf{40.7} & \textbf{11.3} & \textbf{11.3} & \textbf{50.0} \\
\bottomrule
\end{tabular*}
\end{threeparttable}
\end{table}

Replacing \move with atomic click/drag and converting the demonstrations accordingly removes intermediate decision points, reducing overall success from 50.0\% to 19.6\%. Exocentric Drag remains at 100.0\%, but both Ten-Choice variants fall to 2.0\%. Retaining \move with only one supervised move per free or held movement phase yields 27.6\% overall success; the full model improves on all six environments.

The data-ablation model also exhibits \emph{multi-step exploration} in exocentric Ten-Choice: 46 of 61 successful episodes (75.4\%) use two to ten moves, although all 4,000 demonstrations use one move followed by a click. An example trajectory switches candidates after a nonmatching tooltip and clicks only upon finding the target. This feedback-conditioned exploration extends beyond the single-move demonstrations provided for this task.

\FloatBarrier
\subsection{Compatibility with Standard GUI Grounding}
\label{sec:grounding-analysis}
We extend \move to standard GUI grounding using 33,353 selected GroundCUA~\cite{feizi2025groundcua} trajectories, represented as direct clicks or cursor-correction sequences (Section~\ref{sec:move-data}). Intermediate observations display the updated cursor over the same underlying GUI screenshot, allowing the agent to refine its target prediction before clicking.

Using Qwen3-VL-8B as the base model, we compare conventional one-step grounding with \move under both SFT and subsequent RL, alongside the pretrained model. We evaluate all five settings on ScreenSpot-Pro, ScreenSpot-v2, MMBench-GUI, UI-Vision, and OSWorld-G. The two training paths use the same selected grounding records and differ in their action representation and intermediate observations; Appendix~\ref{app:training-details} gives the training settings.

\begin{table}[!ht]
\centering
\small
\setlength{\tabcolsep}{4.5pt}
\begin{tabular}{lcccccc}
\toprule
\textbf{Model} &
\makecell{\textbf{ScreenSpot-}\\\textbf{Pro}} &
\makecell{\textbf{ScreenSpot-}\\\textbf{v2}} &
\makecell{\textbf{MMBench-}\\\textbf{GUI}} &
\textbf{UI-Vision} &
\textbf{OSWorld-G} &
\textbf{Average} \\
\midrule
Qwen3-VL-8B
  & 55.98 & 94.42 & 83.67 & 27.65 & 65.29 & 65.40 \\
Qwen3-VL-8B + SFT
  & 54.84 & 94.26 & 83.31 & 27.76 & 65.29 & 65.09 \\
Qwen3-VL-8B + \move (SFT)
  & 55.72 & 94.18 & 83.11 & 28.13 & 62.94 & 64.82 \\
Qwen3-VL-8B + RL
  & 56.42 & 94.73 & 84.59 & \textbf{31.19} & 66.67 & 66.72 \\
Qwen3-VL-8B + \move (RL)
  & \textbf{58.06} & \textbf{95.28} & \textbf{84.89} & 30.37 & \textbf{68.24} & \textbf{67.37} \\
\bottomrule

\end{tabular}
\caption{Standard GUI grounding accuracy (\%) on five benchmarks. All models use the Qwen3-VL-8B backbone.
``Qwen3-VL-8B + SFT'' uses conventional one-step grounding data, while Qwen3-VL-8B + \move (SFT)
uses synthetic cursor-correction trajectories derived from
GroundCUA records and serialized in the \move format.
``Qwen3-VL-8B + \move (RL)'' further optimizes the SFT model with GRPO
on grounding tasks. Average is the unweighted mean across the five benchmarks.
Bold values denote the best result in each column.}
\label{tab:standard-grounding}
\end{table}

\FloatBarrier
Table~\ref{tab:standard-grounding} shows similar average accuracy after SFT: 64.82\% for \move and 65.09\% for conventional grounding. After RL, \move reaches 67.37\%, exceeding conventional grounding RL at 66.72\% and the pretrained model at 65.40\%. It improves over conventional RL on four of five benchmarks, including ScreenSpot-Pro (58.06\% versus 56.42\%) and OSWorld-G (68.24\% versus 66.67\%); UI-Vision favors conventional RL (31.19\% versus 30.37\%). These results extend the benefit of intermediate visual feedback to cursor correction in standard grounding.

\section{Conclusion}
We introduce \lgw to evaluate whether GUI agents can
acquire task-relevant information through interaction and use it to guide
subsequent control.
Its six matched tasks distinguish available context, revealed semantics,
and action--effect evidence, exposing substantial gaps in current agents.
The \move interface makes intermediate feedback available before
commitment, while training enables models to use this feedback
for exploration and control.
Ablations establish the importance of action granularity
and iterative demonstrations.
Joint RL supports competence across paired control regimes,
and the same interaction format improves standard GUI grounding after RL.
Beyond success rates, feedback-conditioned multi-step exploration
and gain compensation provide behavioral evidence that the trained models
can learn about the environment through interaction and use what they learn
to adjust subsequent actions.
RL further strengthens this adaptive control, as reflected in
stronger gain compensation in egocentric Drag.

\begin{center}
\textbf{A GUI agent should not only learn where to move. It should learn from the move.}
\end{center}

\bibliographystyle{unsrtnat}
\bibliography{references}

\appendix
\section{Benchmark Construction Details}
\label{app:benchmark-construction}

The fixed test suite contains 150 paired cases per environment, each instantiated in both variants, yielding 900 evaluation episodes at $1280\times720$ resolution. Evaluation covers episode-specific configurations. Table~\ref{tab:benchmark-splits} distinguishes variation in layouts and control parameters from explicit holdout of shape categories, message content, or background images. Held-out factors are defined relative to the task-training data.

\begin{table}[ht]
\centering
\small
\setlength{\tabcolsep}{4pt}
\begin{tabularx}{\linewidth}{@{}lcXX@{}}
\toprule
\textbf{Environment} & \makecell{\textbf{Test}\\\textbf{episodes}} & \textbf{Episode-varying factors} & \textbf{Explicitly held-out factors} \\
\midrule
Drag & 150 & Object/slot geometry, appearance, and layout; egocentric sensitivity & Shape categories in 75 episodes; the other 75 use training shape categories \\
Ten-Choice & 150 & Candidate positions and message assignments; egocentric sensitivity & Candidate message content \\
Rotation & 150 & Background image, initial angle, sensitivity, and rotation direction & Background images \\
\bottomrule
\end{tabularx}
\caption{Fixed test-set composition and generalization factors. Counts are per variant; each environment has two paired variants. Episode-specific configuration variation and explicit factor holdout are reported separately.}
\label{tab:benchmark-splits}
\end{table}

\paragraph{Validation set and model selection.}
A separate validation set contains 150 episodes per variant, totaling 900 episodes, matching the test-set size. Validation episodes are disjoint from the training and test episodes. Checkpoint selection uses validation success rates exclusively; the selected checkpoints are then evaluated on the test set for the reported benchmark results.

\paragraph{Drag sampling and shape holdout.}
Each variant has 75 test episodes using training shape categories and 75 using held-out shape categories. The former use six categories for movable objects and matching slots: circle, square, triangle, hexagon, wide rectangle, and cross. The latter use four disjoint categories: pentagon, trapezoid, star, and L-shape. Both subsets vary scene geometry and layout, covering 5--8 candidate slots and four layout families: scattered, jittered grid, ring/arc, and two clusters. The evaluated egocentric variant uses equal positive gains on both axes, with sensitivity drawn from $[0.5,0.8]\cup[1.2,1.5]$, also used in training. Both shape subsets vary episode-specific configurations; shape-category holdout and within-range control-parameter variation are distinct evaluation factors.

\paragraph{Ten-Choice sampling and splits.}
The 150 test scenes use a pool of 240 distinct messages disjoint from the 80 messages in the task-training data. Each scene specifies ten candidate messages, their positions, and a target, all shared by its two variants. The egocentric variant uses equal positive axis gains, with 75 test cases in each sensitivity band, $[0.5,0.8]$ and $[1.2,1.5]$. Training uses the same bands. Thus, all test episodes use held-out message content, while episode-specific gain variation remains within the training support.

\paragraph{Rotation sampling and splits.}
The 150 test pairs use 150 distinct background images, disjoint from the 1,000 backgrounds in the task-training set. Initial angular misalignment, sensitivity, and rotation direction vary across cases. Within each pair, the image, initial relative angle, alignment target, and underlying challenge parameters are shared; the responsive control region differs. This split evaluates held-out image content under episode-varying control parameters.

\paragraph{Episode parameterization and train--test separation.}
Semantic assignments and dynamics parameters are assigned during dataset construction and remain fixed throughout each trajectory. Resetting a fixed test case restores its assigned configuration rather than drawing a new latent. The egocentric Drag and Ten-Choice test variants use positive gains; Rotation varies both response sensitivity and direction. Test cases use separate seeds from task-training cases. Drag additionally separates source cases, Ten-Choice separates message pools, and Rotation separates background images.

\paragraph{Pairing and initial observations.}
Pairing preserves scene content and task objectives. Initial-observation matching follows the construction of each variant. Drag pairs share object and slot identities and world positions; 100 pairs share the initial layout, while 50 place the egocentric target initially offscreen (25 using training shape categories and 25 using held-out shape categories). Ten-Choice pairs share candidate layouts, message assignments, and targets; Rotation pairs share the image and initial alignment challenge. With a common mouse marker, all Ten-Choice and Rotation pairs and the 100 shared-layout Drag pairs have identical first frames. These counts distinguish scene pairing from strict visual matching. Cross-variant matching is distinct from the within-variant latent ambiguity used to define environment-identification levels.

\section{Training Details}
\label{app:training-details}

\subsection{Benchmark SFT}
\label{app:benchmark-training}
LatentLearner-9B initializes from Qwen3.5-9B and is fine-tuned on 24,000 successful trajectories, with 4,000 from each environment. These trajectories yield 154,713 supervised examples. Each example contains the task instruction, up to the latest three screenshots, and preceding assistant actions; the target is a Think rationale followed by one primitive action. The loss supervises the demonstrated response, while the preceding context serves as input.

Training uses VERL with FSDP2, gradient checkpointing, and a maximum sequence length of 16,384 tokens. The configured budget is three epochs; Table~\ref{tab:training-hparams} lists the optimization settings alongside those of the grounding extension. We select one of 30 checkpoints by overall success rate on the separate six-environment validation set (150 episodes per environment, 900 in total) and use it for all six test environments. The benchmark model in Table~\ref{tab:main} receives no RL updates. Benchmark paired RL and grounding RL are detailed separately below.

\subsection{Main-Result Policy Prompt and Conversation Format}
\label{app:main-result-prompt}

This section specifies the policy prompt used for the six-environment main results. LatentLearner-9B uses the same task prompt, action schemas, and history layout during supervised training and evaluation. The public Qwen3.5-9B row uses this evaluation prompt without project fine-tuning.

\paragraph{Environment task instructions.}
Table~\ref{tab:task-instructions} gives the exact text substituted for \texttt{\{task\_instruction\}}. Both variants in each task family use the same instruction. In Ten-Choice, \texttt{\{target\_label\}} is replaced by the episode's requested message. These task instructions are separate from model-specific tool instructions.

\begin{table}[!ht]
\caption{Task instructions for the six evaluation environments. Quotation marks around the Ten-Choice target are part of the instruction.}
\label{tab:task-instructions}
\centering
\small
\begin{tabularx}{\linewidth}{@{}l>{\raggedright\arraybackslash}X@{}}
\toprule
\textbf{Environment} & \textbf{Task instruction} \\
\midrule
Exocentric Drag & \texttt{Place the solid colored shape into the matching gray outline.} \\
Egocentric Drag & \texttt{Place the solid colored shape into the matching gray outline.} \\
Exocentric Ten-Choice & \texttt{Click the icon that displays "\{target\_label\}".} \\
Egocentric Ten-Choice & \texttt{Click the icon that displays "\{target\_label\}".} \\
Inner-ring Rotation & \texttt{Drag the slider to complete verification.} \\
Outer-ring Rotation & \texttt{Drag the slider to complete verification.} \\
\bottomrule
\end{tabularx}
\end{table}
\FloatBarrier

\paragraph{Prompt-ablation suffix.}
For the prompted condition in Section~\ref{sec:prompt-ablation}, GPT-6 Astra receives its baseline task instruction followed by the two sentences below, separated by blank lines. The baseline condition contains neither added sentence; model-specific tool instructions remain unchanged.
\begin{Verbatim}[fontsize=\footnotesize,breaklines=true,frame=single]
Before attempting to solve the task, first explore the environment.

The current environment is egocentric.
\end{Verbatim}

\paragraph{Complete task prompt.}
The following text begins the user message at every decision. The placeholder \texttt{\{task\_instruction\}} is replaced by the episode's task instruction. The six action schemas below are exposed together; each response contains one selected action. The template is shared across the six variants, with the task instruction, observations, and history supplying the episode-specific input.

\begin{Verbatim}[fontsize=\footnotesize,breaklines=true,breakanywhere=true,frame=single]
Task requirement: {task_instruction}
Action space for this task:
Allowed action kinds: move_to, mouse_down, mouse_up, left_click, drag, click
Action schemas:
- move_to: {"kind":"move_to","x":<integer 0-1000>,"y":<integer 0-1000>}
- mouse_down: {"kind":"mouse_down"}
- mouse_up: {"kind":"mouse_up"}
- left_click: {"kind":"left_click"}
- drag: {"kind":"drag","points":[[<integer 0-1000>,<integer 0-1000>],[<integer 0-1000>,<integer 0-1000>]]}
- click: {"kind":"click","points":[[<integer 0-1000>,<integer 0-1000>]]}
Choose exactly one next action from the current screenshot and causal history.
Coordinates must be integers in the 0-1000 screenshot-relative system: x increases left to right and y top to bottom.
Put one or two short English sentences grounded only in visible evidence and causal history inside <think>...</think>, at most 60 words.
Then output one compact JSON object: {"action":{...}}.
Do not mention hidden targets, sensitivities, oracle values, ground-truth actions, formatting, or validation.
Do not add Markdown or any other text.
\end{Verbatim}

\paragraph{Multimodal input and history.}
We construct one user message containing the task prompt followed by image and text blocks. At decision $t$ (counting from one), we retain up to the three most recent pre-action screenshots, including the current observation $O_t$. Every preceding assistant response $R_1,\ldots,R_{t-1}$ is included verbatim, preserving both its Think block and action JSON. Responses whose screenshots have fallen outside the three-image window appear first as text; the retained screenshots are then interleaved with their corresponding responses in chronological order. Screenshots are supplied as image blocks, not as textual scene descriptions. No separate benchmark-specific system message is added; model-native chat serialization is applied by the corresponding processor.

The first decision contains the complete task prompt, the initial screenshot, and this suffix:
\begin{Verbatim}[fontsize=\footnotesize,breaklines=true,frame=single]
No previous actions are present in this trajectory.
The last attached image is the current observation.
Return the Think block and JSON action for step 1.
\end{Verbatim}

For example, the fourth decision is assembled as follows. Bracketed entries denote inserted content rather than literal prompt text; all other lines reproduce the history labels and suffix.
\begin{Verbatim}[fontsize=\footnotesize,breaklines=true,frame=single]
[Complete task prompt above]
Previous step 1 assistant response (context only):
[R_1, verbatim]
[Image O_2]
Previous step 2 assistant response (context only):
[R_2, verbatim]
[Image O_3]
Previous step 3 assistant response (context only):
[R_3, verbatim]
[Image O_4: current observation]
3 previous assistant response(s) are included in chronological order.
The last attached image is the current observation.
Return the Think block and JSON action for step 4.
\end{Verbatim}

\paragraph{Training targets and evaluation responses.}
During SFT, the user message is followed by the demonstrated assistant response, which is the supervised target; earlier responses embedded in the user message provide context. During evaluation, the assistant response is generated from this user message, its action is executed, and the next observation updates the context. Thus, training history contains demonstration responses and evaluation history contains the model's own preceding responses. The response format is a \texttt{<think>...</think>} block followed by one compact JSON object with the top-level key \texttt{action}. The prompt requests one or two short English reasoning sentences, at most 60 words, grounded in visible evidence and prior interaction. Hidden environment parameters, target annotations, and evaluator-only state are not supplied as additional input fields.

\subsection{Benchmark Paired RL Training and Model Selection}
\label{app:paired-rl-selection}
Each paired model starts from the mixed-SFT LatentLearner-9B checkpoint and is trained with GRPO on 5,120 prompts, comprising 2,560 from each variant. The combined dataset is shuffled and used for one epoch. Each prompt generates five rollouts; advantage normalization groups rollouts from the same prompt and environment instance. Training retains the with-think response format, the latest three screenshots, and at most 12 interaction turns. Table~\ref{tab:paired-rl-hparams} lists the training settings.

The reward combines task success ($1$) and trajectory-format validity ($0.1$). In Drag and Ten-Choice, trajectories terminated by generation or context-budget exhaustion receive zero reward. KL regularization is applied as a separate loss, with no KL term added to the reward. All three runs use BF16 and FSDP2.

\begin{table}[!ht]
\caption{Benchmark paired-RL hyperparameters.}
\label{tab:paired-rl-hparams}
\begin{center}
\small
\setlength{\tabcolsep}{5pt}
\begin{tabular}{lc}
\toprule
\textbf{Hyperparameter} & \textbf{All task families} \\
\midrule
Training prompts (per variant) & 2,560 \\
Training budget (epochs) & 1 \\
Optimizer & AdamW \\
Learning rate & $10^{-6}$ \\
Learning-rate schedule & Cosine \\
Warmup ratio & 0.08 \\
Minimum learning-rate ratio & 0.1 \\
Adam $\beta_1,\beta_2$ & $(0.9,0.95)$ \\
Weight decay & 0.05 \\
Gradient clipping norm & 1.0 \\
Generation batch (prompts) & 16 \\
Rollouts per prompt & 5 \\
Rollouts per generation batch & 80 \\
Optimizer minibatch (action samples) & 80 \\
Micro-batch size per GPU & 1 \\
PPO epochs per batch & 1 \\
PPO clipping ratio & 0.2 \\
KL loss coefficient & 0.005 \\
Entropy loss coefficient & 0 \\
Rollout temperature / top-$p$ & $1.0 / 0.95$ \\
Context window (tokens) & 16,384 \\
Trajectory response budget (tokens) & 8,192 \\
Per-turn generation cap (tokens) & 384 \\
Maximum interaction turns & 12 \\
Retained screenshots & 3 \\
\bottomrule
\end{tabular}
\end{center}
\end{table}

Each turn generates at most 384 tokens, subject to the remaining context and trajectory response budgets and a 32-token safety reserve.

Checkpoint selection uses the separate validation set, with 150 validation episodes per variant, matching the test-set size. Each single-environment specialist in Table~\ref{tab:within-pair} is selected by validation success rate on the variant it is trained on, with ties resolved toward the later checkpoint. The paired Drag and Ten-Choice models are selected by mean validation success rate across the two variants with the same tie rule. Paired Rotation is selected by inner-ring validation success. Each paired model supplies both test results within its family.

\subsection{Grounding SFT and RL}
\label{app:grounding-rejection-sampling}

\paragraph{Training data.}
For grounding SFT, we use 33,353 training trajectories, including 30,018 one-move, 1,668 two-move, and 1,667 three-move trajectories, giving an approximately $90{:}5{:}5$ distribution. The conventional direct-click baseline and the \move variant use the same selected records with different action representations. Subsequent RL uses 10,000 grounding instructions excluded from the SFT instruction set.

\paragraph{Rejection sampling for coordinate supervision.}
We construct coordinate supervision by generating and verifying candidate grounding actions with Qwen3-VL-8B-Instruct. The model receives the task instruction, screenshots, and preceding actions, while the annotated target box remains hidden and is used only for offline verification. Predicted coordinates are integers in $[0,1000]$ on each axis and are mapped back to the original image for the target-box check. A coordinate is accepted as the terminal target when it falls inside the annotated box, including its boundary. Malformed actions and out-of-range coordinates trigger retries.

If the first valid prediction misses the target, its coordinate is retained as an intermediate cursor position. We render the cursor at that position on the unchanged screenshot and request a correction conditioned on the updated observation and action history. The second valid prediction is checked in the same way. If both predictions miss, or format retries are exhausted, the pipeline supplies a recovery coordinate that lies inside the target box after coordinate quantization. Recovery coordinates are recorded separately from model predictions. Valid misses can therefore remain as intermediate correction steps, while every completed trajectory ends at a verified target position.

\paragraph{Trajectory construction and serialization.}
For the \move variant, coordinate-bearing clicks are converted into \texttt{move\_to} actions without changing their coordinates, followed by a coordinate-free \texttt{left\_click}. Intermediate observations update only the cursor overlay, while the underlying GUI screenshot remains fixed. During rebalancing, some three-move trajectories are shortened by removing the second miss while preserving the first miss and the successful endpoint. We then select training trajectories according to the length distribution described above.

The direct-click baseline retains only the final coordinate-bearing click for each selected record. The \move variant preserves intermediate actions and observations and uses training windows containing at most three screenshots together with the preceding textual action history. Windowing yields 35,020 training examples from the 33,353 trajectories. Actions already supervised in an earlier window are masked from the loss in subsequent windows. Both settings use action-only targets without Think annotations or assistant prefilling.

\paragraph{Supervised fine-tuning.}
Both SFT settings initialize from Qwen3-VL-8B-Instruct and use full-parameter fine-tuning with VERL and FSDP2. Training uses BF16 precision, FlashAttention 2, gradient checkpointing, and random seed 1. The direct-click baseline uses 32 GPUs with four gradient-accumulation steps, while the \move variant uses 16 GPUs with eight accumulation steps. Both settings maintain a global batch size of 128 and train for two epochs. Optimization settings are summarized in Table~\ref{tab:training-hparams}.

\paragraph{Reinforcement learning.}
RL initializes from the corresponding SFT checkpoint and uses GRPO with eight rollouts per prompt. Rewards are $-0.2$ for invalid format, $0$ for a valid but incorrect result, and $1$ for a valid and correct result. We apply a KL regularization loss with coefficient $0.01$, without adding KL to the reward. The \move policy receives updated cursor observations between actions, with at most four actions and three retained screenshots; the direct-click policy predicts one coordinate-bearing click. Rejection sampling constructs the offline SFT supervision, whereas RL optimizes trajectories newly generated by the current policy.

\begin{table}[ht]
\centering
\small
\setlength{\tabcolsep}{5pt}
\begin{tabular}{lccc}
\toprule
\textbf{Hyperparameter} & \textbf{Benchmark SFT} & \textbf{Grounding SFT} & \textbf{Grounding RL} \\
\midrule
Training budget (epochs) & 3 & 2 & 1 \\
Learning rate & $3\times10^{-6}$ & $3\times10^{-6}$ & $5\times10^{-7}$ \\
Learning-rate schedule & Cosine & Cosine & Cosine \\
Warmup ratio & 0.08 & 0.05 & 0.05 \\
Weight decay & 0.05 & 0.1 & 0.01 \\
Gradient clipping norm & 1.0 & 1.0 & 1.0 \\
Training batch size & 64 examples & 128 examples & 8 prompts \\
Micro-batch size per GPU & 1 & 1 & 1 \\
Maximum sequence length & 16,384 & 32,768 & 16,384 \\
Rollouts per prompt & -- & -- & 8 \\
Rollout temperature & -- & -- & 1.0 \\
Rollout top-$p$ & -- & -- & 0.95 \\
KL loss coefficient & -- & -- & 0.01 \\
\bottomrule
\end{tabular}
\caption{Training hyperparameters. Benchmark SFT uses a three-epoch budget, with the reported checkpoint selected by overall validation success rate. Grounding RL uses GRPO on 10,000 instructions excluded from grounding SFT.}
\label{tab:training-hparams}
\end{table}

\clearpage
\section{Frontier Evaluation Protocols}
\label{app:frontier-protocols}

\paragraph{GPT-6 Astra.}
We evaluate GPT-6 Astra through a persistent \texttt{exec\_py} session that supports mouse operations and screenshot capture. Each episode allows up to 12 model responses; actions within a response are executed until the environment terminates.

\paragraph{GPT-5.6-Sol.}
We evaluate GPT-5.6-Sol through the native \texttt{computer} tool. Exocentric Drag and both Ten-Choice variants use its computer actions. For egocentric Drag and both Rotation variants, the interface additionally exposes independent \texttt{mouse\_down} and \texttt{mouse\_up} functions for controlling the held-button state. Native computer actions provide screenshot feedback, while the button functions return completion messages. Each episode allows up to 12 model responses; all actions in a response are executed unless the environment terminates.

\paragraph{Claude Opus 5.}
We evaluate Claude Opus 5 through the native computer-use interface, retaining the structured conversation history and screenshots. Each episode allows up to 12 model responses, including screenshot requests. All actions in a response are executed until the environment terminates. Rotation is scored at the first release attempt.

\paragraph{Kimi-K3.}
We evaluate Kimi-K3 through screenshot-based tool calling with pixel-coordinate actions. The local executor supports cursor movement, button presses and releases, clicks, and drags. Across all six environments, Kimi-K3 uses the 13 KimiCU MCP tools together with four additional tools: \texttt{move\_to}, \texttt{mouse\_down}, \texttt{mouse\_up}, and \texttt{left\_click}. The adapter provides an initial screenshot and an updated screenshot after each action, and the prompt instructs the agent to inspect feedback while holding a drag. Each episode allows up to 12 model calls, with one tool call per response. The conversation history preserves reasoning and screenshots. Episodes ending in protocol errors count as failures.

\clearpage
\section{Gain-Dependent Control after SFT and Paired RL}
\label{app:mixed-sft-behavior}
We analyze the egocentric Drag evaluation of the jointly trained Qwen3.5-9B SFT policy reported in Table~\ref{tab:main}. The evaluation contains 150 complete episodes, including 62 successes.

\paragraph{Measurement.}
For each \texttt{move\_to} action that produces nonzero held-object movement, let $u_t$ be the commanded displacement in screen pixels and $e_t$ the pre-action target-minus-object residual, also in screen pixels. We measure the error-normalized command component along the residual,
\[
\alpha_t=\frac{u_t^{\mathsf T}e_t}{e_t^{\mathsf T}e_t},\qquad
\bar\alpha_i=\operatorname{median}_{t\in\mathcal{M}_i}\alpha_t,
\]
where $\mathcal{M}_i$ contains the eligible movements in episode $i$. Pearson correlations are computed across episodes between $\bar\alpha_i$ and either hidden gain $g_i$ or its inverse $1/g_i$, giving each episode equal weight. Under the evaluated positive isotropic gain law, ideal unclamped correction has $\alpha=1/g$. Hidden gain and geometry are read from evaluator-only logs for offline measurement and are not supplied to the policy.

\paragraph{Coverage and robustness.}
We apply no terminal-success filter. The full movement cohort contains 155 actions from 80 episodes: 62 successful and 18 failed, with 30 low-gain ($g<1$) and 50 high-gain ($g>1$) episodes. The remaining 70 episodes contain no held-object movement and all fail; their movement coefficient is undefined. The correlation therefore describes episodes with observed held-object movement.

A filtered cohort retains movements with a visible target, residual norm at least 20 pixels, an ideal correction within the action-coordinate range, and no clamping. We recompute the episode medians after filtering. Table~\ref{tab:mixed-sft-gain} reports both correlations for the full and filtered cohorts of each policy. Correlations with $1/g$ describe alignment with ideal compensatory scaling; filtered results check robustness.

\begin{table}[!ht]
\caption{Gain-dependent control after mixed SFT and paired RL in egocentric Drag. Pearson correlations relate episode-median normalized command to hidden gain $g$ and ideal compensation $1/g$, using the same aggregation and movement filters for both policies.}
\label{tab:mixed-sft-gain}
\begin{center}
\small
\setlength{\tabcolsep}{5pt}
\begin{tabular}{llrrrr}
\toprule
Policy & Cohort & Episodes & Moves & $r(\bar\alpha,g)$ & $r(\bar\alpha,1/g)$ \\
\midrule
Mixed SFT & All held moves & 80 & 155 & $-0.592$ & $0.610$ \\
 & Filtered & 72 & 85 & $-0.557$ & $0.616$ \\
\midrule
Paired RL & All held moves & 145 & 270 & $-0.820$ & $0.803$ \\
 & Filtered & 135 & 177 & $-0.749$ & $0.745$ \\
\bottomrule
\end{tabular}

\end{center}
\end{table}

For mixed SFT, median $\bar\alpha$ is 1.232 in the low-gain band and 0.755 in the high-gain band; after filtering, the corresponding values are 1.375 and 0.755. For the full cohort, median ideal $1/g$ is 1.528 in the low-gain band and 0.729 in the high-gain band. The association reflects compensatory command scaling across the two gain bands.

\paragraph{Comparison with paired RL.}
The paired-RL evaluation succeeds on 138/150 episodes (92.0\%), compared with 62/150 (41.3\%) for mixed SFT. Two episodes with protocol errors are included in these counts. Its 145 movement episodes comprise 138 successes and seven failures; the other five episodes have no held movement and fail.

On the 79 episode IDs with held movement under both policies, correlations between $\bar\alpha$ and gain $g$ are $-0.591$ for SFT and $-0.813$ for paired RL (154 and 143 moves, respectively). The stronger association therefore persists on shared episode IDs.

\clearpage
\section{Paired Policy Case Studies}
\label{app:paired-policy-cases}

We select two successful test pairs from the paired-RL policies in Table~\ref{tab:within-pair}. Within each pair, the validation-selected policy, task-prompt text, initial screenshot pixels, and first action are identical. All frames and actions come from recorded model evaluations. The selected cases illustrate how feedback guides subsequent actions; Table~\ref{tab:within-pair} reports success across the full test set.

\subsection{Ten-Choice: Semantic Reveal and Egocentric Control}
\label{app:paired-ten-case}

In pair \texttt{ed-ten-0010}, both runs seek ``This is the jade obelisk.'' The shared first move reveals a nonmatching tooltip under $I_1$ but shifts the candidate field without revealing a tooltip under $I_2$. The next actions differ accordingly. Both runs continue searching and click only after the target message is visible, succeeding in eight and five actions, respectively.

\begin{figure}[!ht]
\centering
{\small\bfseries Shared initial observation; shared first action: MOVE(94,758)}\par\smallskip
\includegraphics[width=.22\linewidth]{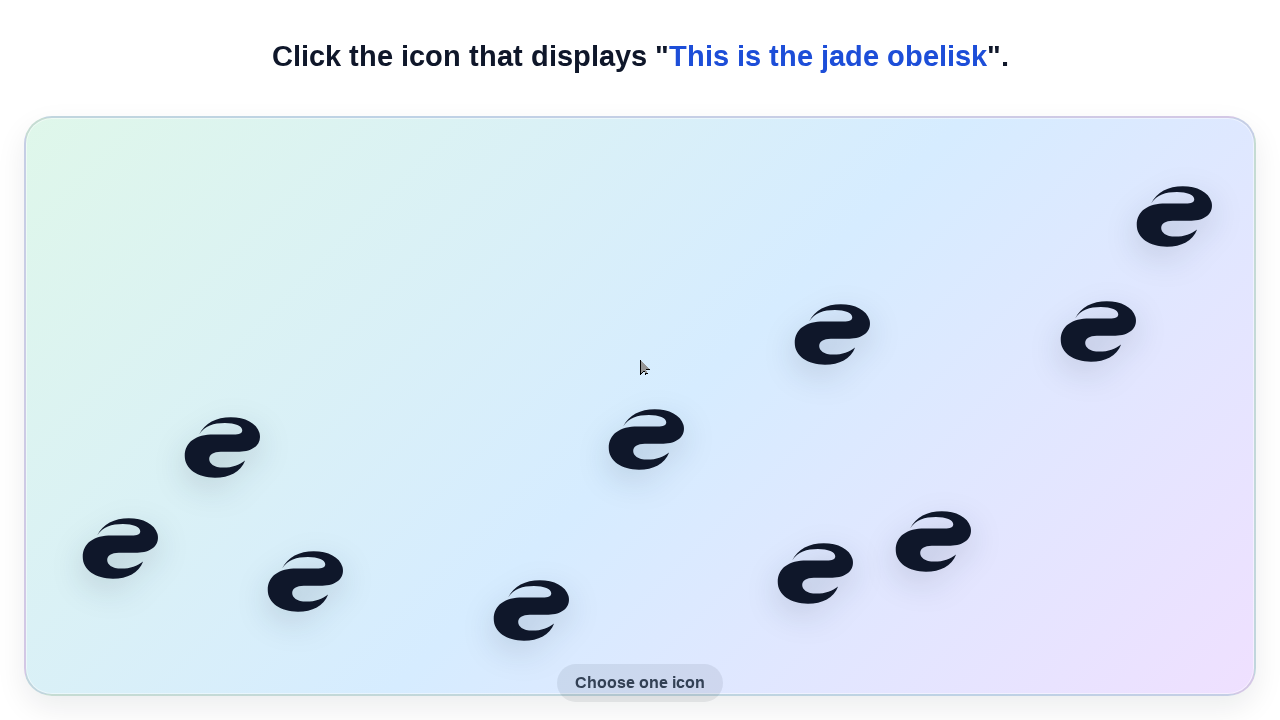}\par\smallskip
\begin{minipage}[t]{.48\linewidth}
\centering
{\small\bfseries Exocentric: semantic reveal ($I_1$)}\par\smallskip
\includegraphics[width=.78\linewidth]{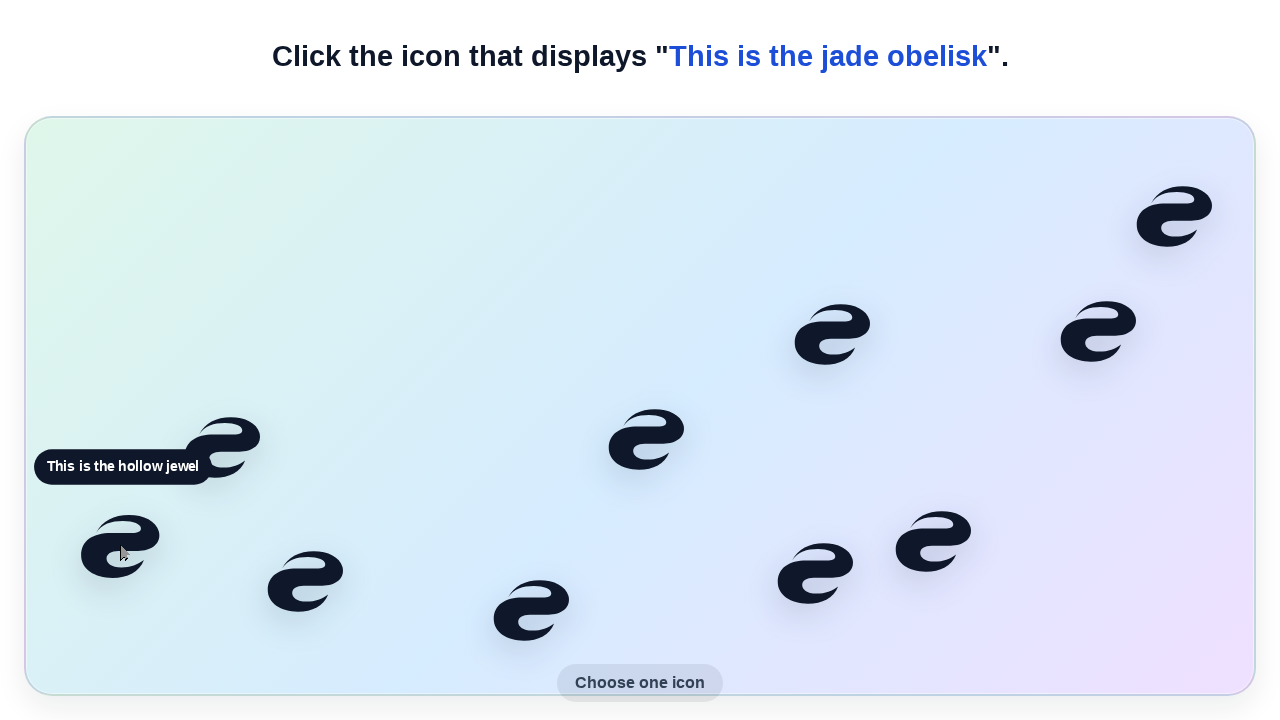}\par
{\footnotesize After action 1: ``This is the hollow jewel''}\par
{\footnotesize Next: MOVE(173,618)}\par\medskip
\includegraphics[width=.78\linewidth]{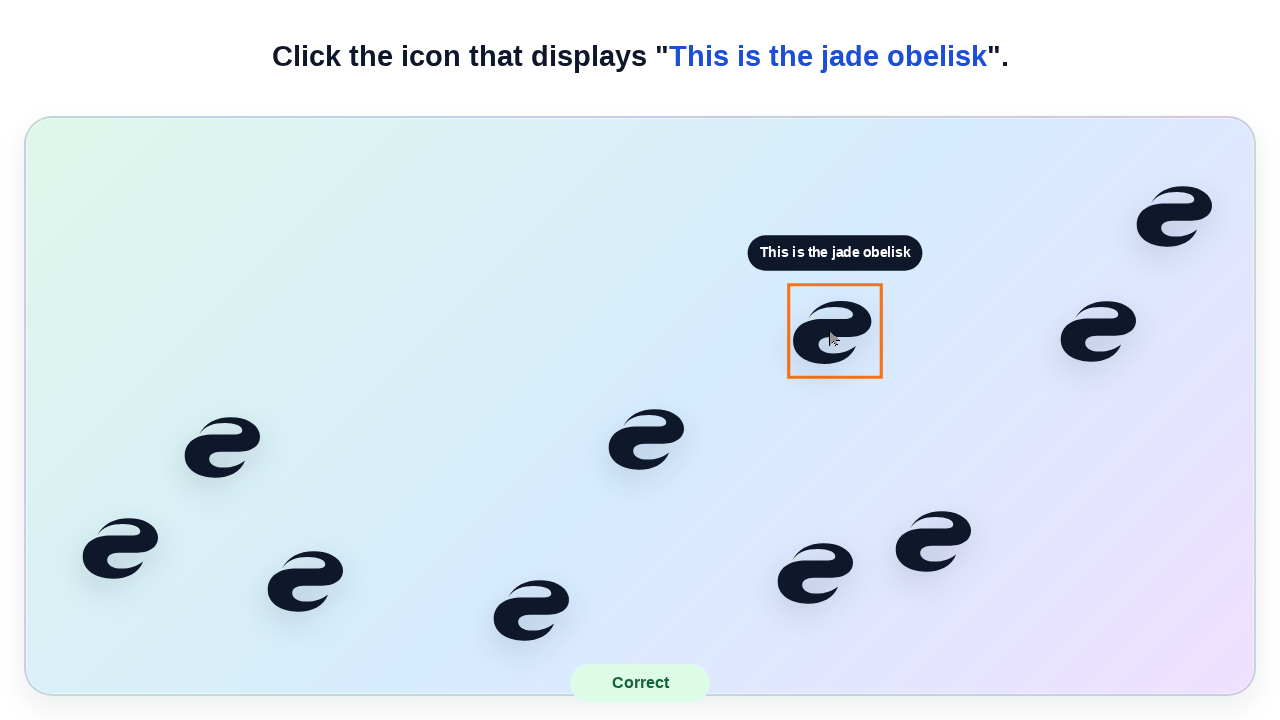}\par
{\footnotesize Success after action 8: target selected}
\end{minipage}\hfill
\begin{minipage}[t]{.48\linewidth}
\centering
{\small\bfseries Egocentric: scene control ($I_2$)}\par\smallskip
\includegraphics[width=.78\linewidth]{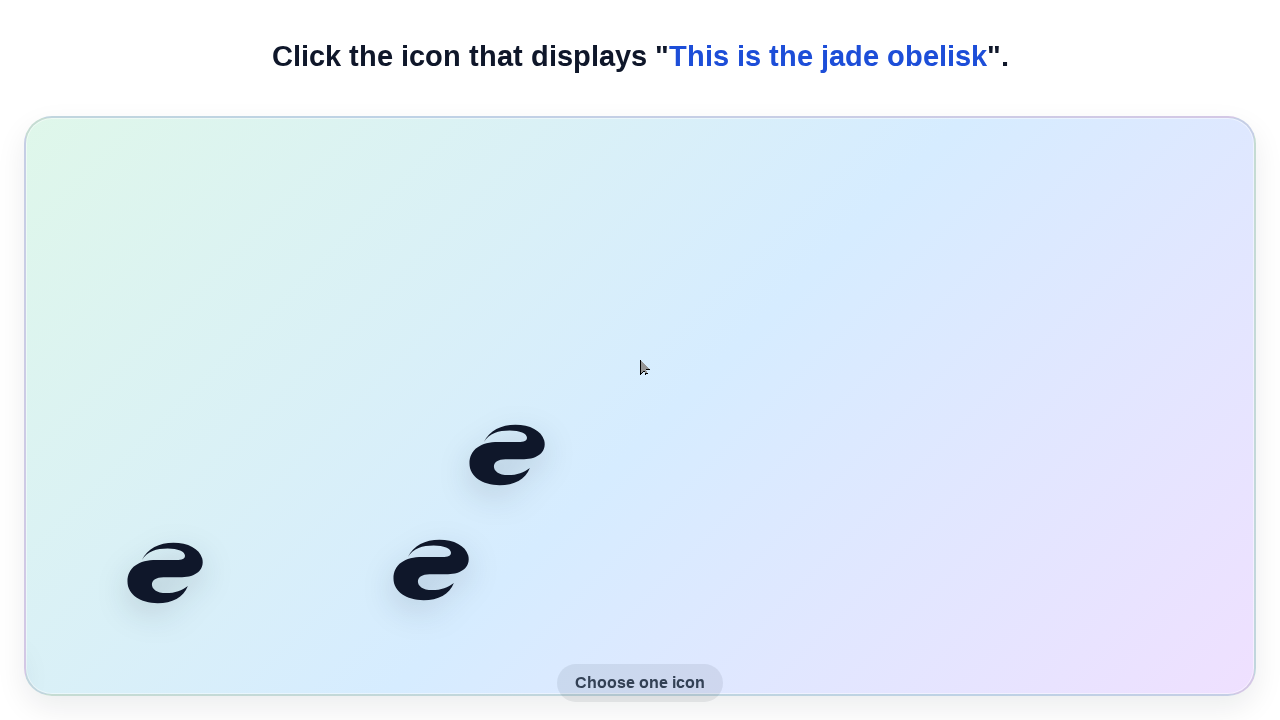}\par
{\footnotesize After action 1: scene shifted; no tooltip}\par
{\footnotesize Next: MOVE(575,400)}\par\medskip
\includegraphics[width=.78\linewidth]{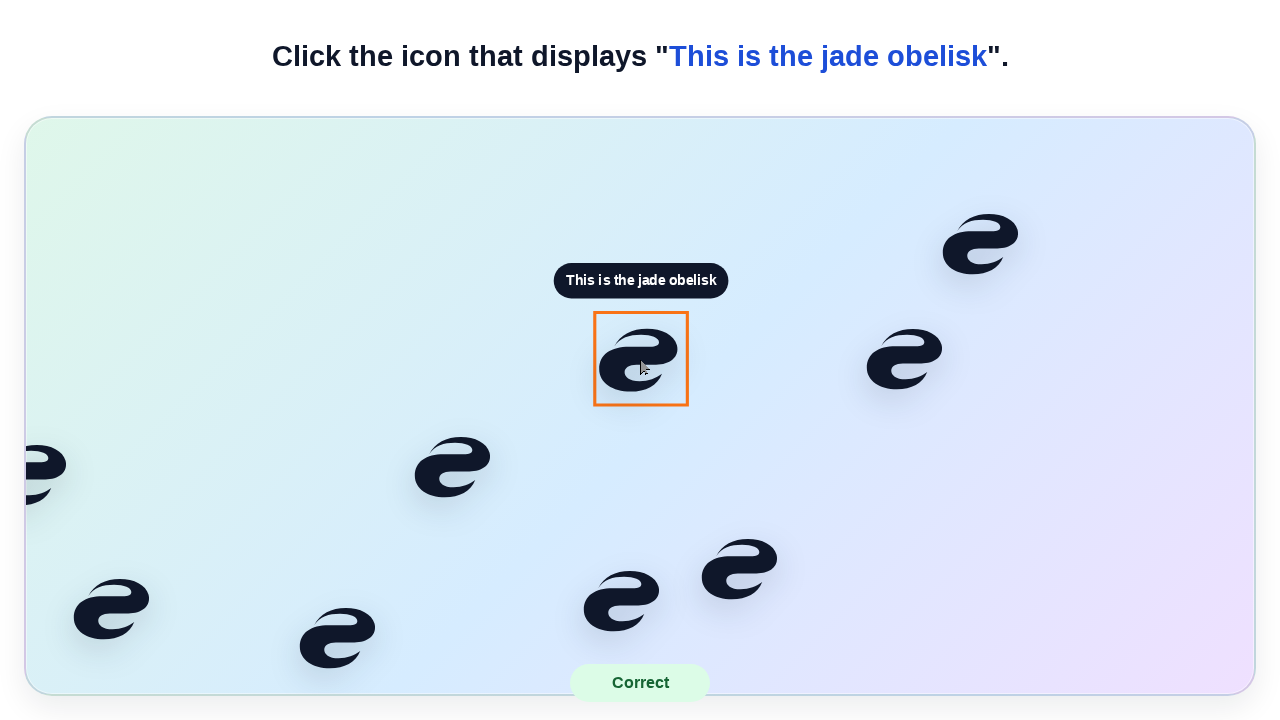}\par
{\footnotesize Success after action 5: target selected}
\end{minipage}
\par\medskip
{\footnotesize
\begin{tabularx}{\linewidth}{@{}lXX@{}}
\toprule
& Exocentric ($I_1$) & Egocentric ($I_2$) \\
\midrule
Actions & $M(94,758)\to M(173,618)$\newline
$\to M(238,804)\to M(414,844)$\newline
$\to M(503,607)\to M(635,793)$\newline
$\to M(648,460)\to C$
& $M(94,758)\to M(575,400)$\newline
$\to M(551,385)\to M(662,487)$\newline
$\to C$ \\
\bottomrule
\end{tabularx}
}
\caption{\textbf{Different feedback after the same Ten-Choice action.} The cursor moves over fixed candidates on the left; candidates move around the fixed cursor on the right. Native screenshots and complete actions are shown. $M$ denotes \texttt{MOVE}, $C$ denotes a current-position click; coordinates are normalized to $[0,1000]$.}
\label{fig:paired-ten-policy}
\end{figure}

\clearpage
\subsection{Drag: Direct Placement and Egocentric Correction}
\label{app:paired-drag-case}

In pair \texttt{ed-drag-0004}, both runs begin with \texttt{MOVE(403,828)} toward the visible blue rectangle. Under $I_0$, the cursor arrives on the object, and the next action grasps it. Under $I_2$, the scene shifts and the object remains offset from the fixed cursor; the policy first issues \texttt{MOVE(522,433)} and only then grasps. While holding the object, it makes two observed movements before release, compared with one in the direct-control variant. Both runs succeed on their first release, using four and six actions, respectively. The case illustrates feedback-dependent acquisition and correction under the two mechanisms.

\begin{figure}[!ht]
\centering
{\small\bfseries Shared initial observation; shared first action: MOVE(403,828)}\par\smallskip
\includegraphics[width=.26\linewidth]{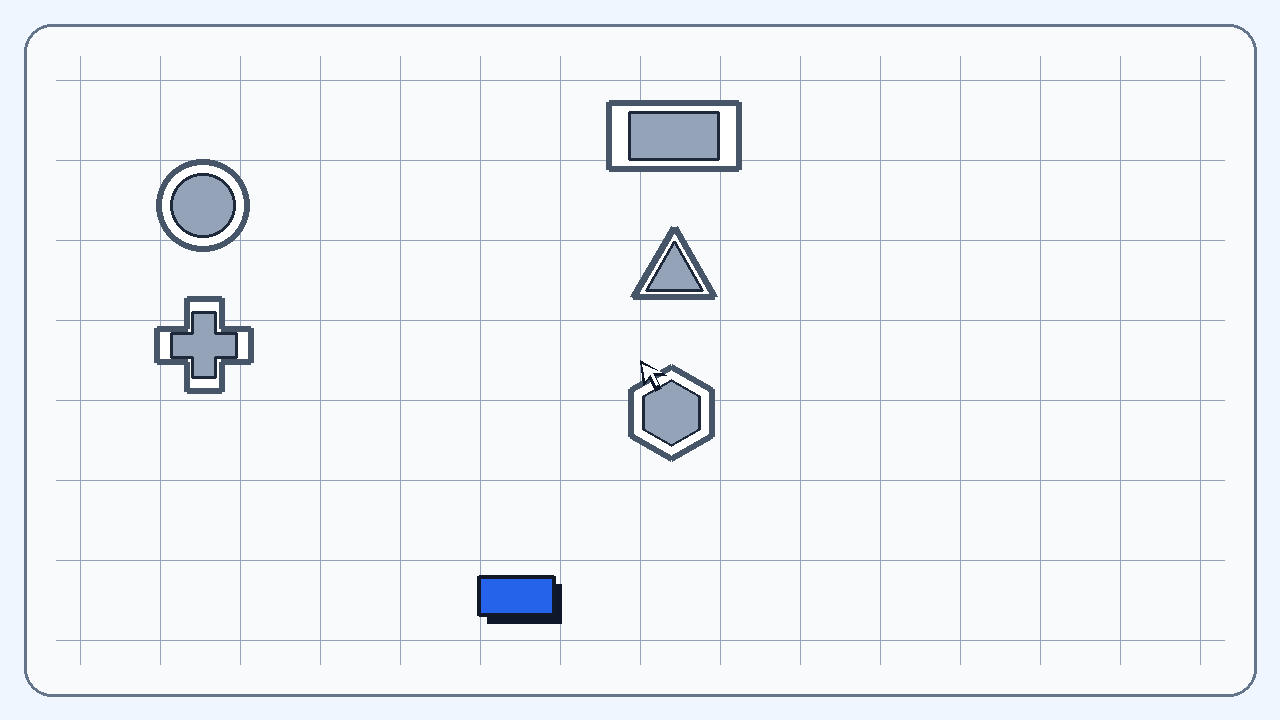}\par\smallskip
\begin{minipage}[t]{.48\linewidth}
\centering
{\small\bfseries Exocentric: direct control ($I_0$)}\par\smallskip
\includegraphics[width=.82\linewidth]{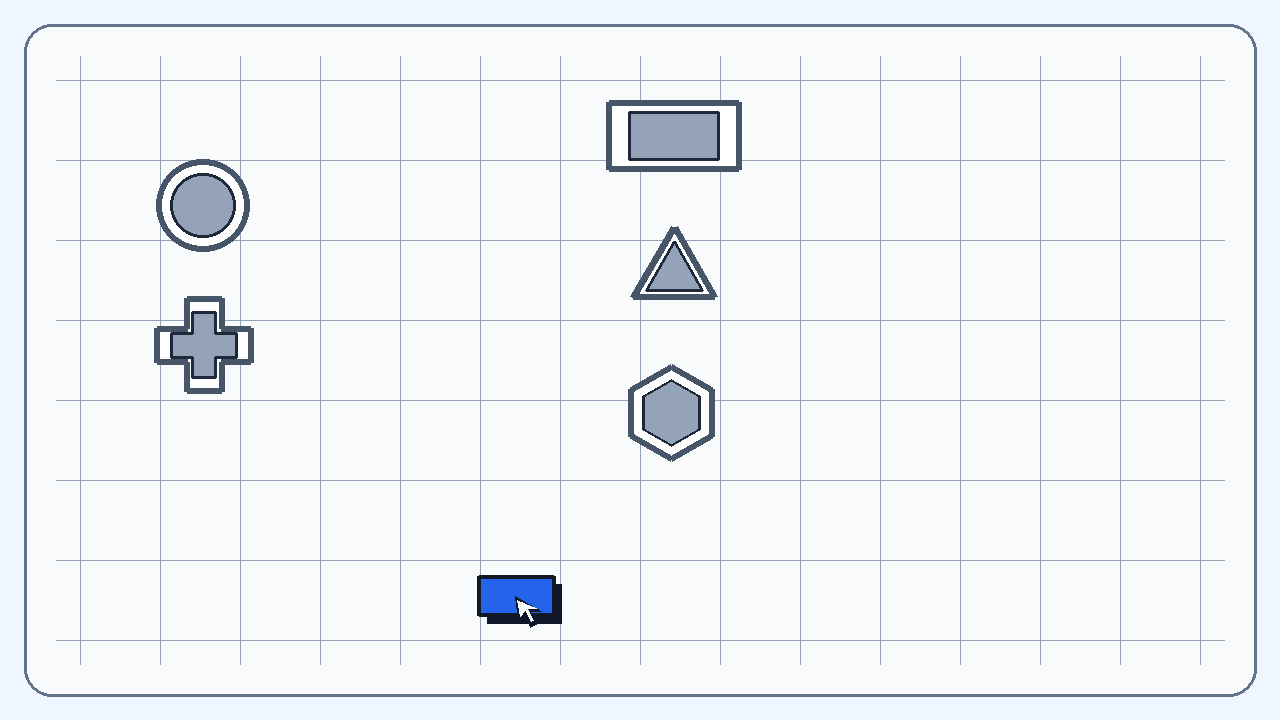}\par
{\footnotesize After action 1: cursor on the blue rectangle}\par
{\footnotesize Next: MOUSE\_DOWN}\par\medskip
\includegraphics[width=.82\linewidth]{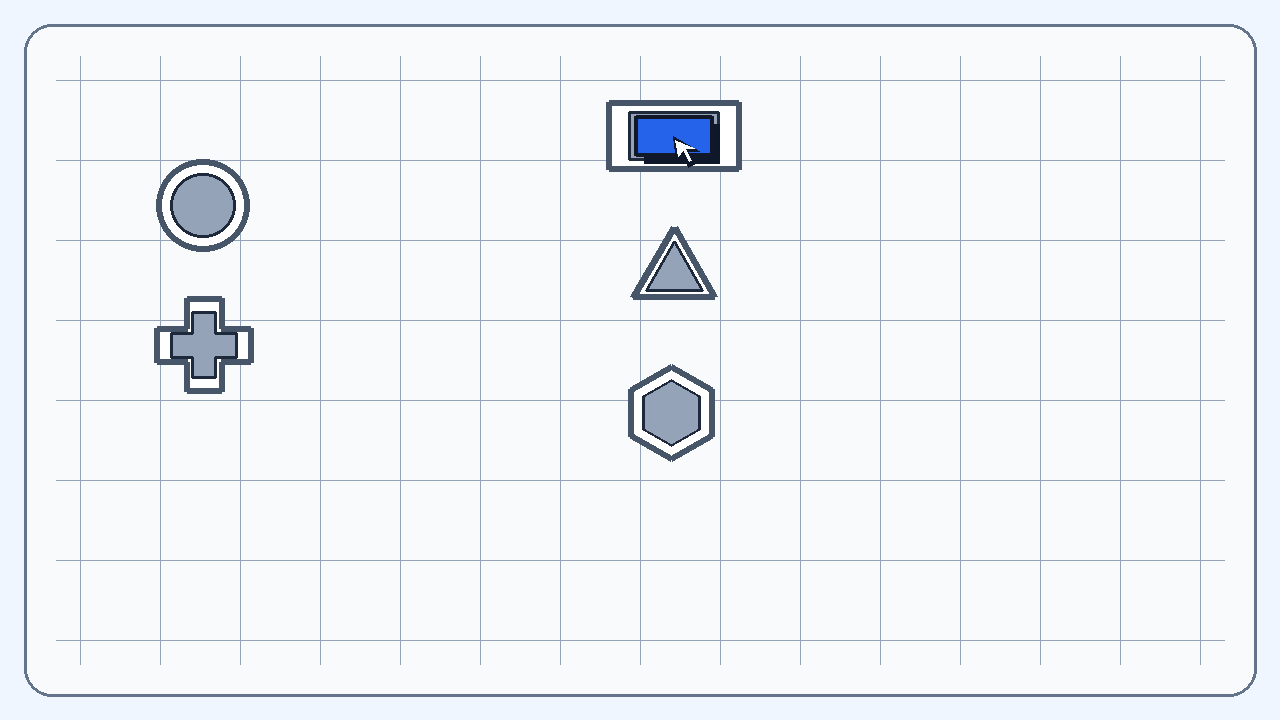}\par
{\footnotesize Success after action 4: first release}
\end{minipage}\hfill
\begin{minipage}[t]{.48\linewidth}
\centering
{\small\bfseries Egocentric: viewpoint control ($I_2$)}\par\smallskip
\includegraphics[width=.82\linewidth]{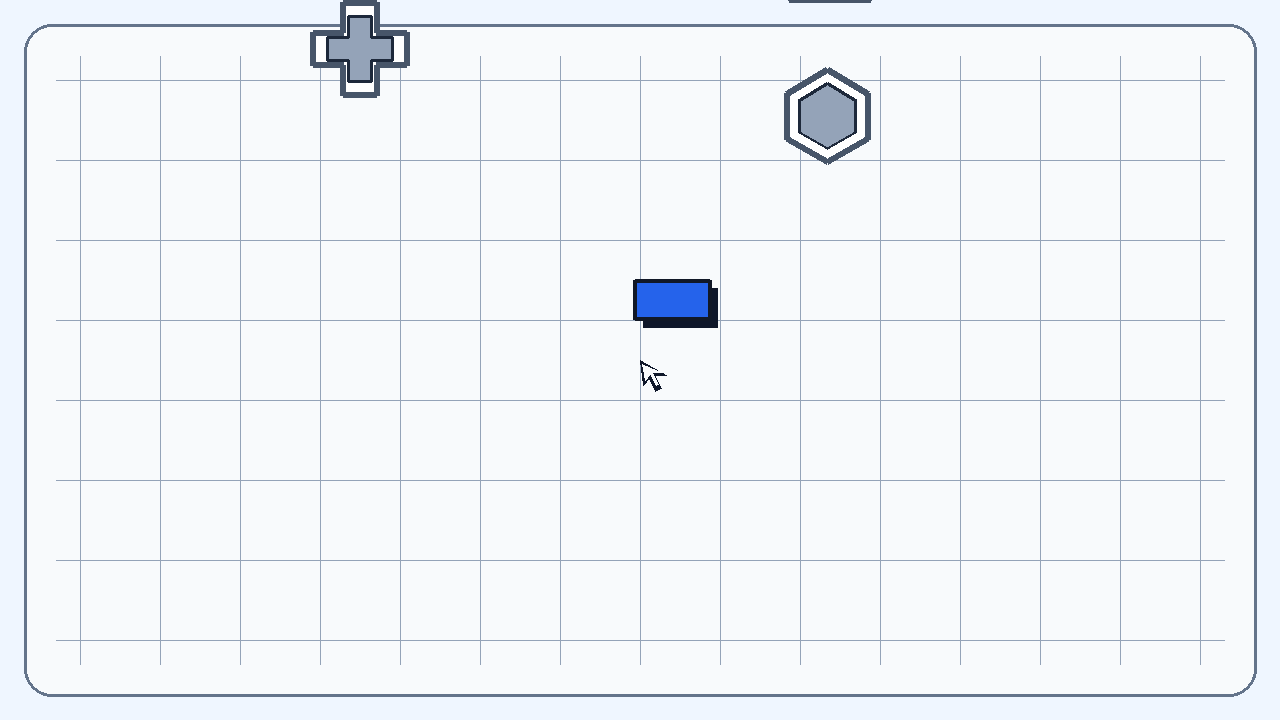}\par
{\footnotesize After action 1: object offset from the cursor}\par
{\footnotesize Next: MOVE(522,433)}\par\medskip
\includegraphics[width=.82\linewidth]{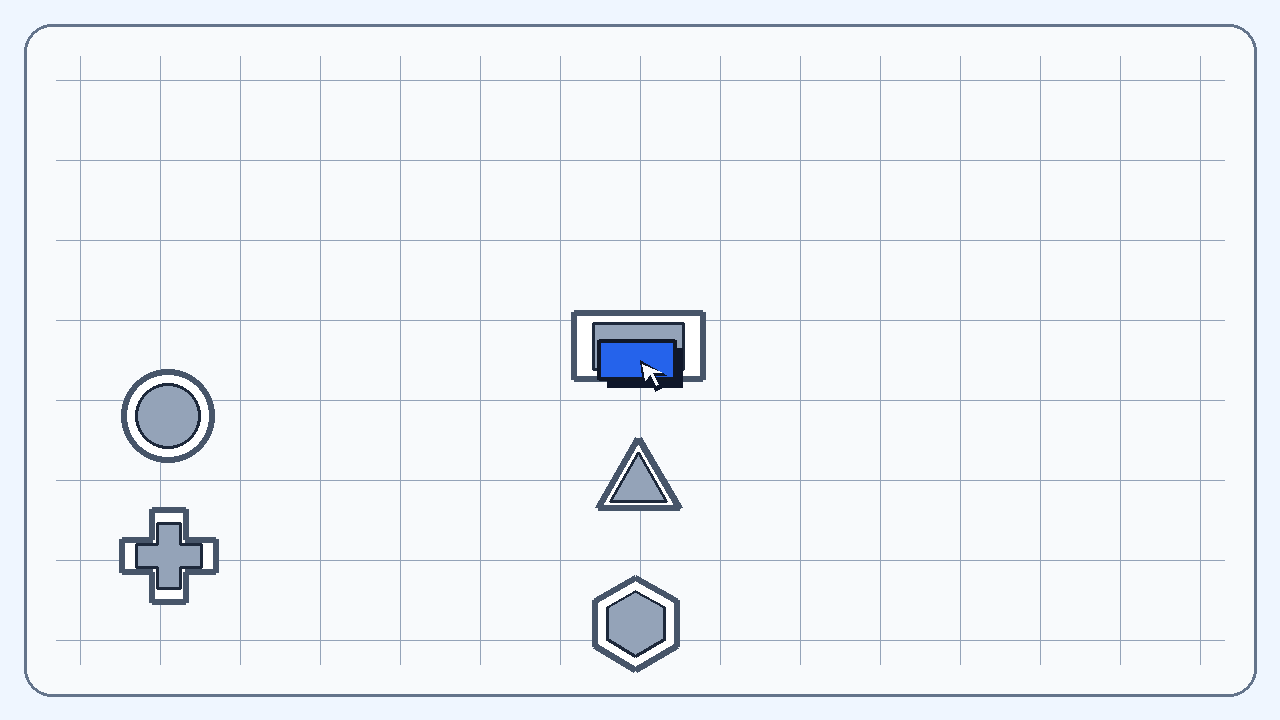}\par
{\footnotesize Success after action 6: first release}
\end{minipage}
\caption{\textbf{Feedback changes when the Drag policy grasps and releases.} The initial scene and first action are identical, but the next action is a press under direct control and another move under egocentric control. Both trajectories end in successful placement. Native screenshots are shown unchanged; Table~\ref{tab:paired-drag-actions} lists every recorded action.}
\label{fig:paired-drag-policy}
\end{figure}

\begin{table}[!ht]
\centering
\small
\begin{tabularx}{\linewidth}{@{}lXX@{}}
\toprule
& Exocentric ($I_0$) & Egocentric ($I_2$) \\
\midrule
Actions & $M(403,828)\to D$\newline
$\to M(526,189)\to U$
& $M(403,828)\to M(522,433)\to D$\newline
$\to M(600,166)\to M(497,340)\to U$ \\
Outcome & Success; 4 actions & Success; 6 actions \\
\bottomrule
\end{tabularx}
\caption{Complete recorded action sequences for the Drag pair. $M$ denotes \texttt{MOVE}, $D$ denotes \texttt{MOUSE\_DOWN}, and $U$ denotes \texttt{MOUSE\_UP}. Coordinates are normalized to $[0,1000]$.}
\label{tab:paired-drag-actions}
\end{table}

\end{document}